\def\year{2015}
\documentclass[letterpaper]{article}
\usepackage{aaai}

\usepackage{times}
\usepackage{helvet}
\usepackage{courier}
\usepackage{url}
\usepackage{epsfig}
\usepackage{graphicx}
\usepackage{amsmath}
\usepackage{amssymb}
\usepackage{wrapfig}
\usepackage{subfigure}
\usepackage{amsmath,amsthm,amssymb,mathrsfs,amsfonts}
\usepackage{algorithm2e, algorithmic}
\RestyleAlgo{boxruled}

\usepackage{appendix}
\usepackage{chngcntr}

\usepackage{color}
\usepackage{pdflscape}

\newcommand{\etal}{\textit{et al}.}
\newcommand{\ie}{\textit{i}.\textit{e}.}
\newcommand{\eg}{\textit{e}.\textit{g}.}
\newcommand{\bLambda}{\boldsymbol{\Lambda}}
\newcommand{\bU}{\boldsymbol{U}}
\newcommand{\bV}{\boldsymbol{V}}

\newcommand{\bx}{\boldsymbol{x}}
\newcommand{\by}{\boldsymbol{y}}
\newcommand{\bI}{\boldsymbol{I}}
\newcommand{\bK}{\boldsymbol{K}}
\newcommand{\bM}{\boldsymbol{M}}
\newcommand{\bX}{\boldsymbol{X}}
\newcommand{\bL}{\boldsymbol{L}}

\newcommand{\inner}[2]{\left\langle#1,#2\right\rangle}

\newtheorem{prop}{Proposition}
\newenvironment{numberedprop}[1]
  {\renewcommand\theprop{#1}\prop}
  {\endprop}

\nocopyright
 
\begin{document}

%
\title{Jointly Learning Multiple Measures of Similarities from Triplet Comparisons}
\author{ {\bf Liwen Zhang} \\
University of Chicago\\
liwenz@cs.uchicago.edu\\
\And
{\bf Subhransu Maji}  \\
UMass Amherst         \\
smaji@cs.umass.edu\\
\And
{\bf Ryota Tomioka}   \\
Toyota Technological Institue at Chicago \\
tomioka@ttic.edu \\
}

\nocopyright

\maketitle
\begin{abstract}
\begin{quote}
Similarity between objects is multi-faceted and it can be easier for
human annotators to measure it when the focus is on a specific aspect. 
We consider the problem of mapping objects into view-specific embeddings where the distance between them is consistent with the similarity comparisons of the form ``from the t-th view, object A is more similar to B than to C''. 
Our framework \emph{jointly} learns view-specific embeddings exploiting correlations between views. 
Experiments on a number of datasets, including one of multi-view
 crowdsourced comparison on bird images, show the proposed method
 achieves lower triplet generalization error when compared to both learning
 embeddings independently for each view and all views pooled into one view.
Our method can also be used to learn multiple measures of similarity
 over input features taking class labels into account and compares favorably to existing approaches for multi-task metric learning on the ISOLET dataset. 
\end{quote}
\end{abstract}

\section{Introduction}

Measure of similarity plays an important role in applications such as
content-based recommendation, image search and speech
recognition. Therefore a  number of techniques to {\em learn} a measure
of similarity from data have been
proposed~\cite{xing2002distance,DavKulJaiSraDhi07,WeiBliSau06,mcfee2011learning}. 
When the measure of distance is induced by an inner product in a low-dimensional
space as is done in many studies, learning a distance metric is
equivalent to learning an {\em embedding} of objects in a low-dimensional space.
This is useful for visualization 
 as well as using the learned representation in a variety of
down-stream tasks that require fixed length representations of
 objects as has been demonstrated by the applications of word embeddings~\cite{mikolov2013efficient} in language.

Among various forms of supervision for learning distance metric,
 similarity comparison of the form 
`object $A$ is more similar to $B$ than to $C$'', which we call {\em
triplet comparison}, is extremely useful for obtaining 
an embedding that reflects a {\em perceptual similarity}
\cite{agarwal2007generalized,tamuz2011adaptively,van2012stochastic}. Triplet
comparisons can be obtained by crowdsourcing, or
it may also be derived from class labels if available. 

\begin{figure}[tb]
\begin{center}
\includegraphics[width=0.8\linewidth]{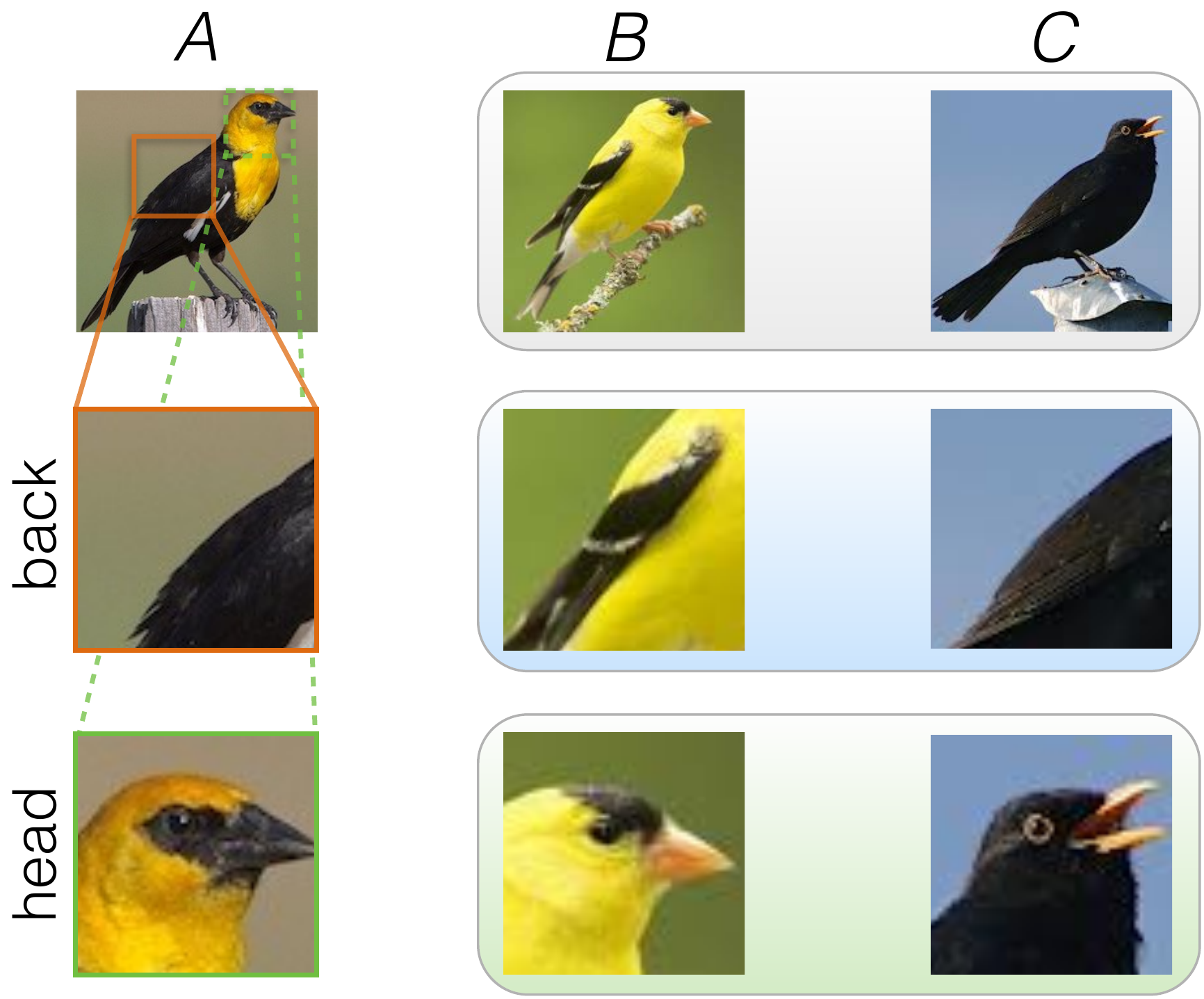}
\caption{\textbf{Ambiguity in similarity.} Depending on whether we focus
 on the back (middle row) or on the head (bottom row), bird $A$ may appear
 more similar to $B$ or $C$. \label{fig:figure1}}
\end{center}
\vspace{-0.2in}
\end{figure}

The task of judging similarity comparisons, however,
can be challenging for human annotators.
Consider the problem of comparing three birds as seen in 
Fig.~\ref{fig:figure1}. Most annotators will say that the head of bird
$A$ is more similar to the head of $B$ while the back of $A$ is more
similar to $C$. Such ambiguity leads to noise in annotation and results in poor embeddings.

A better approach would be to tell the annotator the desired view or the
perspective of the object to use for measuring similarity. Such
view-specific comparisons are not only easier for annotators, but they can also enable precise feedback for human ``in the loop'' tasks, such as, interactive fine-grained recognition~\cite{wah15learning}, thereby reducing the human effort. The main drawback of learning view specific embeddings {\em
independently} is that the number of similarity comparisons scales linearly with the number of views. This is undesirable as even learning a single embedding of $N$
objects may require $O(N^3)$ triplet comparisons~\cite{jamieson2011low} in the worst case.

We propose a method for learning embeddings {\em jointly} that addresses
this drawback. Our method exploits underlying correlations that may exist between the views
allowing a better use of the training data. 
%
%
Our method models the correlation between views by assuming that each
view is a \emph{low-rank projection} of a common embedding. Our model
can be seen as a matrix factorization model in which local metric is defined
as $\bL\bM_t\bL^\top$, where $\bL$ is a matrix that parametrizes the
common embedding and $\bM_t$ is a positive semidefinite matrix
parametrizing the individual view. The model can be efficiently trained
by alternately updating the view specific metric and the common embedding.

We experiment with a synthetic dataset and two realistic datasets,
namely, poses of airplanes, and crowd-sourced similarities collected on
different body parts of birds (CUB dataset; Welinder et al., \citeyear{WelinderEtal2010}).
On most datasets our joint learning approach obtains lower triplet generalization error compared to the independent learning approach or naively pooling all the views into a single one,
especially when the number of training triplets is
limited.
Furthermore, we apply our joint metric learning approach to the
multi-task metric learning setting studied by
\cite{parameswaran2010large} to demonstrate that our method can also
take input features and class labels into account. Our method compares
favorably to the previous method on ISOLET dataset.

\section{Formulation\label{sec:formulation}}

In this section, we first review the single view metric learning problem
considered in previous work.  Then we extend it to the case where there are multiple measures of similarity.

\subsection{Metric learning from triplet comparisons}
Given a set of triplets $\mathcal{S}=\{(i,j,k)\mid\text{object $i$ is
  more similar to object $j$ than object $k$}\}$ and possibly input features $\bx_1,\ldots,\bx_N\in \mathbb{R}^H$,
we aim to find a positive semidefinite matrix $\bM\in\mathbb{R}^{H\times H}$
 such that the pair-wise comparison of the distances induced by the
 inner product $\inner{\bx}{\by}_{\bM}=\bx^\top\bM\by$ parametrized by
  $\bM$ (approximately) agrees with $\mathcal{S}$, \ie, $(i,j,k)\in
\mathcal{S}\Rightarrow \|\bx_i- \bx_j\|_{\bM}^2 < \|\bx_i -
 \bx_k\|_{\bM}^2$. If no input feature is given, we take $\bx_i$ as
 the $i$th coordinate vector in $\mathbb{R}^{N}$, and learning $\bM$,
 which would become $N\times N$, would correspond to finding
 {\em embeddings} of the $N$ objects in a Euclidean space with dimension equal
 to the rank of $\bM$.

Mathematically the problem can be expressed as follows:
\begin{align}
\label{eq:gnmds-k}
\min_{\substack{\bM \in \mathbb{R}^{H\times H},\\
\bM \succeq 0}}  \quad & \!\!\!\!\!\sum_{(i,j,k) \in \mathcal{S}}\!\!\!\!\! 
\ell(\| \bx_i - \bx_j \|_{\bM}^2, \|\bx_i - \bx_k\|_{\bM}^2 )
 +\gamma\text{tr}(\bM),
\end{align}
where $\|\bx-\by\|_{\bM}^2=(\bx-\by)^\top\bM(\bx-\by)$; 
 the loss function can be, for example, logistic \cite{CoxMilMinPapYia00}, or
hinge,
$\ell(d_{i,j},d_{i,k})=\max(1+d_{i,j}-d_{i,k},0)$
 \cite{agarwal2007generalized,WeiBliSau06,CheShaShaBen10}. 
Other choices of loss functions lead to crowd kernel
learning~\cite{tamuz2011adaptively}, and $t$-distributed stochastic
triplet embedding (t-STE)~\cite{van2012stochastic}. Penalizing the trace of the
 matrix $\bM$ can be seen as a convex surrogate for penalizing the rank \cite{agarwal2007generalized,FazHinBoy01}. $\gamma>0$ is a 
regularization parameter.

 After the optimal
$\bM$ is obtained, we can find a low-rank factorization of $\bM$ as
$\bM=\bL\bL^\top$ with $\bL\in\mathbb{R}^{H\times D}$. This is particularly useful when no input feature is
provided, because each row of $\bL$, which is $N\times D$ in this case,
corresponds to a $D$ dimensional embedding of each object.




%

\subsection{Jointly learning multiple metrics}
\label{sec:mmte}
Now let's 
assume that $T$ sets of triplets
$\mathcal{S}_1,\ldots,\mathcal{S}_T$ are available. This 
can be obtained by asking annotators to
focus on a specific aspect when making pair-wise comparisons as in 
human in the loop tasks \cite{wah14similarity,wah15learning}.
Alternatively, different measures of similarity can come from
considering multiple related metric learning problems as in
\cite{parameswaran2010large,RaiLiaCar14}.

While a simple approach to handle multiple similarities would be to
parametrize each aspect or view by a positive semidefinite matrix $\bM_t$, this
would not induce any shared structure among the views. Our goal is to
learn a global transformation $\bL$ that maps the objects in a common $D$
dimensional space as well as local view-specific metrics $\bM_t$ ($t=1,\ldots,T$).

To this end, we formulate the learning problem as follows:
\begin{align}
\min_{\substack{ \bL\in\mathbb{R}^{H\times D},\\
\bM_t \in\mathbb{R}^{D\times D},\\
\bM_t \succeq 0\,(t=1,\ldots,T)}} &\!\!\sum_{t=1}^T  \sum_{(i,j,k) \in
 \mathcal{S}_t}\!\!\!\!\!\varphi_{i,j,k}(\bL,\bM_t)\notag \\[-1.2em]
&\quad  +\gamma\sum_{t=1}^T {\rm tr}(\bM_t) + \beta \|\bL\|_F^2 \, , \label{eq:multi-metric-obj} 
\end{align}
where 
$\varphi_{i,j,k}(\bL,\bM):=\ell\bigl(\|\bL^\top(\bx_i-\bx_j)\|_{\bM}^2,\|\bL^\top(\bx_i-\bx_k)\|_{\bM}^2 \bigr)$, and 
$\ell$ is a loss function as above. We use the hinge loss in the
experiments in this paper, but the proposed framework readily generalizes to
other loss functions proposed in literature \cite{tamuz2011adaptively,van2012stochastic}.
Note again that when no input feature is provided, the global
transformation matrix $\bL$ becomes an $N\times D$ matrix that consists
of $D$ dimensional embedding of the objects.

Intuitively the global transformation $\bL$ plays the role of a
bottleneck and forces the local metrics to share the common $D$ dimensional
subspace because they are restricted in the form
$\bL\bM_t\bL^\top$. 


The proposed model \eqref{eq:multi-metric-obj} includes various simpler
models as special cases. First, if $\bL$ is an $H\times H$ identity
matrix, there is no sharing across different views and indeed the
objective function will decompose into a sum of view-wise objectives; we
call this {\em independent learning}. On the other hand, if we constrain
all $\bM_t$ to be equal, the same metric will apply to all the views
and the learned metric will be essentially the same as learning a single
shared metric as in Eq. \eqref{eq:gnmds-k} with
$\mathcal{S}=\cup_{t=1}^{T}\mathcal{S}_t$; we call this {\em pooled learning}.

We employ regularization terms for both the local metric $\bM_t$ and the
global transformation matrix $\bL$ in \eqref{eq:multi-metric-obj}. The
trace penalties ${\rm tr}(\bM_t)$ are employed to obtain low-rank
matrices $\bM_t$ as above.
 The regularization term on
the norm of $\bL$ is necessary to resolve the scale ambiguity. Although
the above formulation has two hyperparameters $\beta$ and $\gamma$, we
show below in Proposition \ref{lem:effective} that the product $\beta\gamma$
is the only hyperparameter that needs to be tuned.

To minimize the objective \eqref{eq:multi-metric-obj}, we update $\bM_t
$'s and $\bL$ alternately. Both updates are (sub)gradient descent. The
$\bM_t$ update is followed by a projection onto the positive
semi-definite (PSD) cone.
Note that 
if we choose a convex loss function, \eg, hinge-loss, then it becomes a
convex problem with respect to $\bM_t$'s 
and $\bM_t$'s can be optimized independently since they appear in disjoint
terms.
The algorithm is summarized
in \textbf{Algorithm~\ref{alg:alt-update}}. 

\subsubsection*{Effective regularization term}
The sum of the two regularization terms employed in
\eqref{eq:multi-metric-obj} can be reduced into a single 
effective regularization term with only {\em one hyperparameter}  $\sqrt{\beta
\gamma}$  as we show in the following proposition (we give the proof in the supplementary material). 

\begin{prop}\label{lem:effective}
\begin{align*}
\!\! \min_{\substack{\bL\in\mathbb{R}^{H\times D},\\
\bM_1,\ldots,\bM_T\in\mathbb{R}^{D\times D}} }
\!\!\!\!&\gamma\sum_{t=1}^{T}
{\rm tr}(\bM_t) + \beta \|\bL\|_F^2
=2\sqrt{
\beta\gamma}{\rm tr}\left(\sum_{t=1}^{T}\bK_t\right)^{\frac{1}{2}},\\
{\rm s.t.}\quad &\bL\bM_t\bL^\top=\bK_t\, (\forall t)
\end{align*}
where the power $1/2$ in the r.h.s. is the matrix square root.
\end{prop}
As a corollary, we can always reduce or maintain the regularization terms in
\eqref{eq:multi-metric-obj} without affecting the loss term by the
rescaling $\bM_t\leftarrow \bM_t/\alpha^2$ and $\bL\leftarrow\alpha\bL$
with $\alpha = ( \gamma \sum_{t=1}^T \operatorname{tr}(\bM_t)/(\beta \|\bL\|_F^2) )^{1/4}$.

\subsubsection*{Number of parameters}
A simple parameter
counting argument tells us that {\em independently} learning $T$ views
requires to fit $O(DHT)$ parameters, where $H$ is the number of input
dimension, which can be as large as $N$, $D$ is the embedding dimension,
and $T$ is the number of views. On the
other hand, our {\em joint learning} model has only $O(HD+D^2T)$
parameters. Thus when $D<H$, our model has much fewer parameters and
enables better generalization, especially when the number of triplets is limited.

\subsubsection*{Efficiency}
Reducing the dimension from $H$ to $D$ by the common transformation
$\bL$ is also favorable in terms
of computational efficiency. The projection of $\bM_t$ to the cone of $D\times
D$ PSD matrices is much more efficient when $D\ll H$ compared to
independently learning $T$ views.


%
%
%

\begin{algorithm}
\label{alg:alt-update}
\begin{small}
 \KwIn{the number of objects $N$ (or input features $(\bx_i)_{i=1}^{N}$); dimension of embedding $D$;
 triplet constraints $\mathcal{S}_t$, $t=1,\ldots,T$;
 regularization parameters  $\beta$, $\gamma$; the
 number of inner gradient updates $m_{\rm max}$} 
 \KwOut{Global transformation $\bL$; PSD matrices $\{{ \bM_t }\}_{t=1}^{T}$}
 Initialize $\bL$ randomly; initialize ${ \bM_t }$ as identity matrices\;
 \While{not converged}{
 Update $\bL$ using step-size $\eta = \eta_0 / \sqrt{m}$ for
 $m_{\rm max}$ times as 
\begin{align*}
  \bL \leftarrow \bL  - \eta \left\{
\sum_{t=1}^{T}\sum_{
\substack{(i,j,k)\in \mathcal{S}_t\\
}}
\nabla_{\bL} \varphi_{i,j,k}(\bL,\bM_t)
+2\beta\bL \right\}
\end{align*}
  \For{ $t \in \{1,2,\dots,T\}$}{
 Update $\bM_t$ using step-size $\eta= \eta_0 / \sqrt{m}$ for
 $m_{\rm max}$ times by taking a gradient step
 \begin{align*}
\!  { \bM_t }  \leftarrow & { \bM_t } - \eta \left\{
\!\! \sum_{(i,j,k)\in \mathcal{S}_t} \!\!\!\!\!\!\nabla_{{ \bM_t }} 
\varphi_{i,j,k}(\bL,\bM_t)  + \gamma \bI_D
\right\}
 \end{align*}
 and projecting ${ \bM_t }$ to the PSD cone\;

  }
 }
  \caption{Multiple-metric Learning}
 \end{small}

\end{algorithm}

\section{Learning embeddings from  triplet comparisons\label{sec:exp}}

In this section, we demonstrate the statistical efficiency of our model
in both the triplet embedding (no input feature), and multi-task metric
learning scenarios (with features).


\subsection{Experimental setup}

\begin{figure}[tb]
\begin{center}
\begin{tabular}{cc}
\includegraphics[width=.38\linewidth]{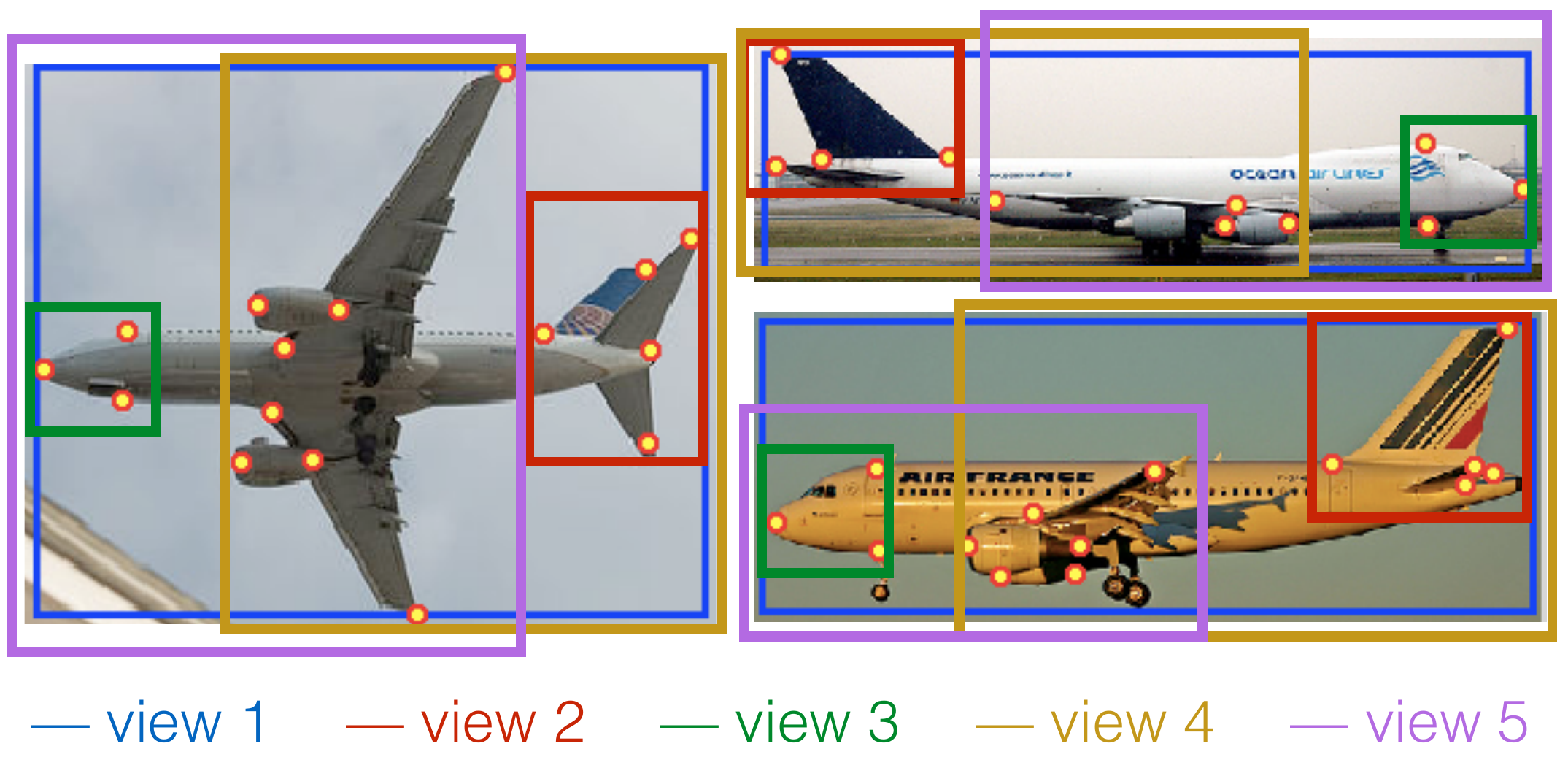} & 
\includegraphics[width=.55\linewidth]{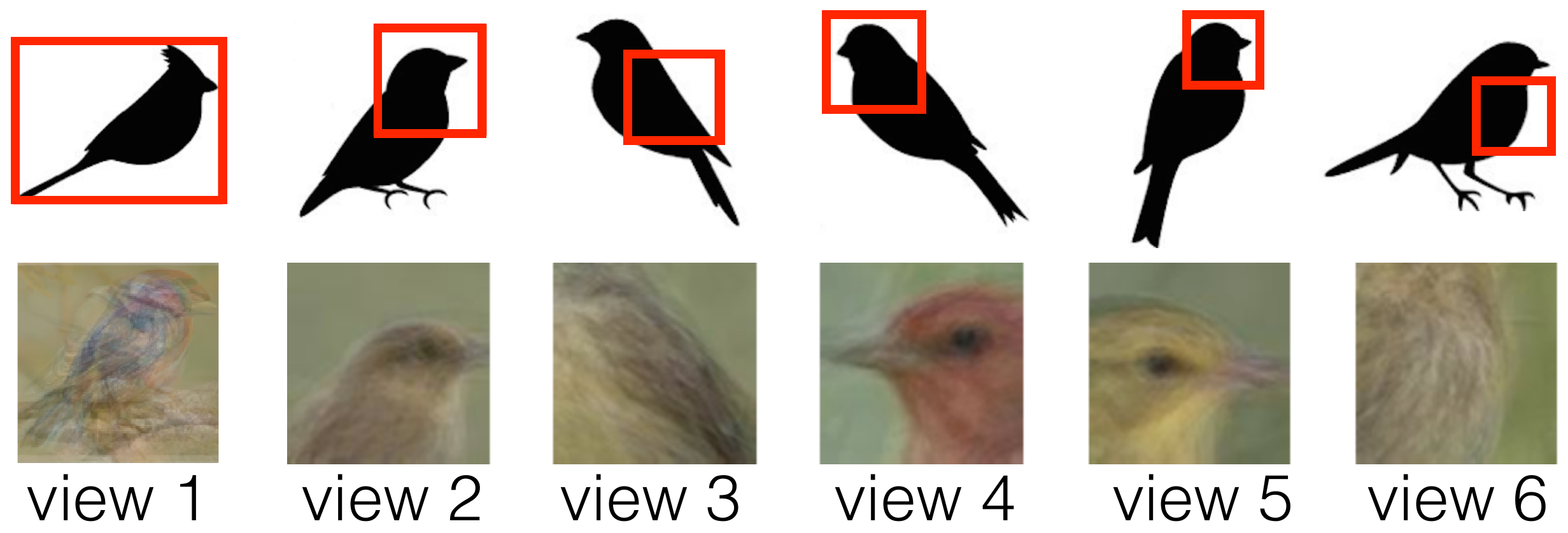}
\end{tabular}
\caption{\label{fig:birds-views} (Left) View-specific similarities between poses of planes were obtained by considering subsets of landmarks shown by different colored rectangles and measuring their similarity in configuration up to a scaling and translation. (Right) Perceptual similarities between bird species were collected by showing users either the full image (view $1$), or crops around various parts (view $2, 3, 4, 5, 6$). The average image for each view is also shown.}
\end{center}
\vspace{-0.1in}
\end{figure}


On each dataset, we divided the triplets into training and test and
measured the quality of embeddings by the triplet generalization error,
i.e., the fraction of test triplets whose relations are incorrectly
modelled by the learned embedding. The error was measured for each view
and averaged across views. The numbers of training triplets were the
same for all the views.
The regularization parameter was tuned using a  
5-fold cross-validation on the training set with candidate values
$\left\{ 10^{-5}, 10^{-4}, \dots , 10^{5}\right\}$. 
The hinge loss was used as the loss function. We use $m_{\rm max}=20$
as the number of inner iterations in the experiments.

In addition, we inspected how the similarity knowledge on existing views
could be {\em transferred} to a new view where the number of similarity
comparisons is small. We did this by conducting an experiment in which
we drew a small set of training triplets from one view but used large
numbers of training triplets from the other views.

We compared our method with the following two baselines. {\bf Independent}: We conducted triplet embedding on each view treating
 each of them independently.  We parametrized $\bM=\bL\bL^\top$ with
 $\bL\in\mathbb{R}^{N\times D}$ and minimized \eqref{eq:gnmds-k}
 using the software provided by van der Maaten and Weinberger~\shortcite{van2012stochastic}.
{\bf Pooled}: We learned a single embedding with 
 the training triplets from all the views combined.

\subsection{Synthetic data}

\paragraph{Description}
Two synthetic datasets were generated.  One consisted of 200 points uniformly sampled from a 10 dimensional unit hypercube, while the other dataset had 200 objects from a mixture of four Gaussian with variance 1 whose centers were randomly chosen in a hypercube with side length 10.  Six \textit{views} were generated on each dataset.  
Each \textit{view} was produced by projecting data points onto a random subspace.  The dimensions of the six random subspaces were 2, 3, 4, 5, 6, and 7 respectively.

\paragraph{Results}

\begin{figure}
\centering
	\subfigure[Clustered]{
    \includegraphics[width=0.3\linewidth]{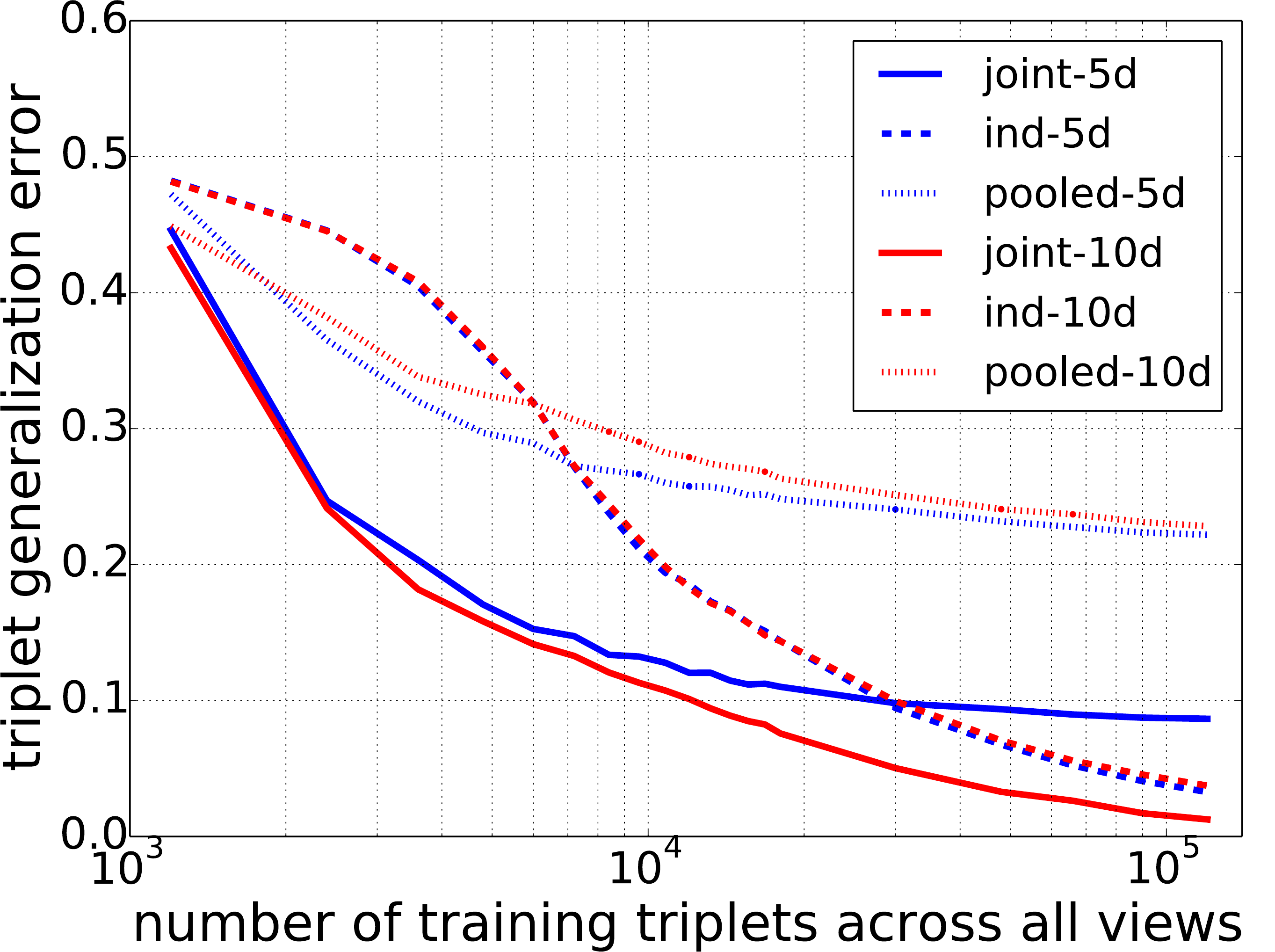}
    }
    \subfigure[Uniform]{
    \includegraphics[width=0.3\linewidth]{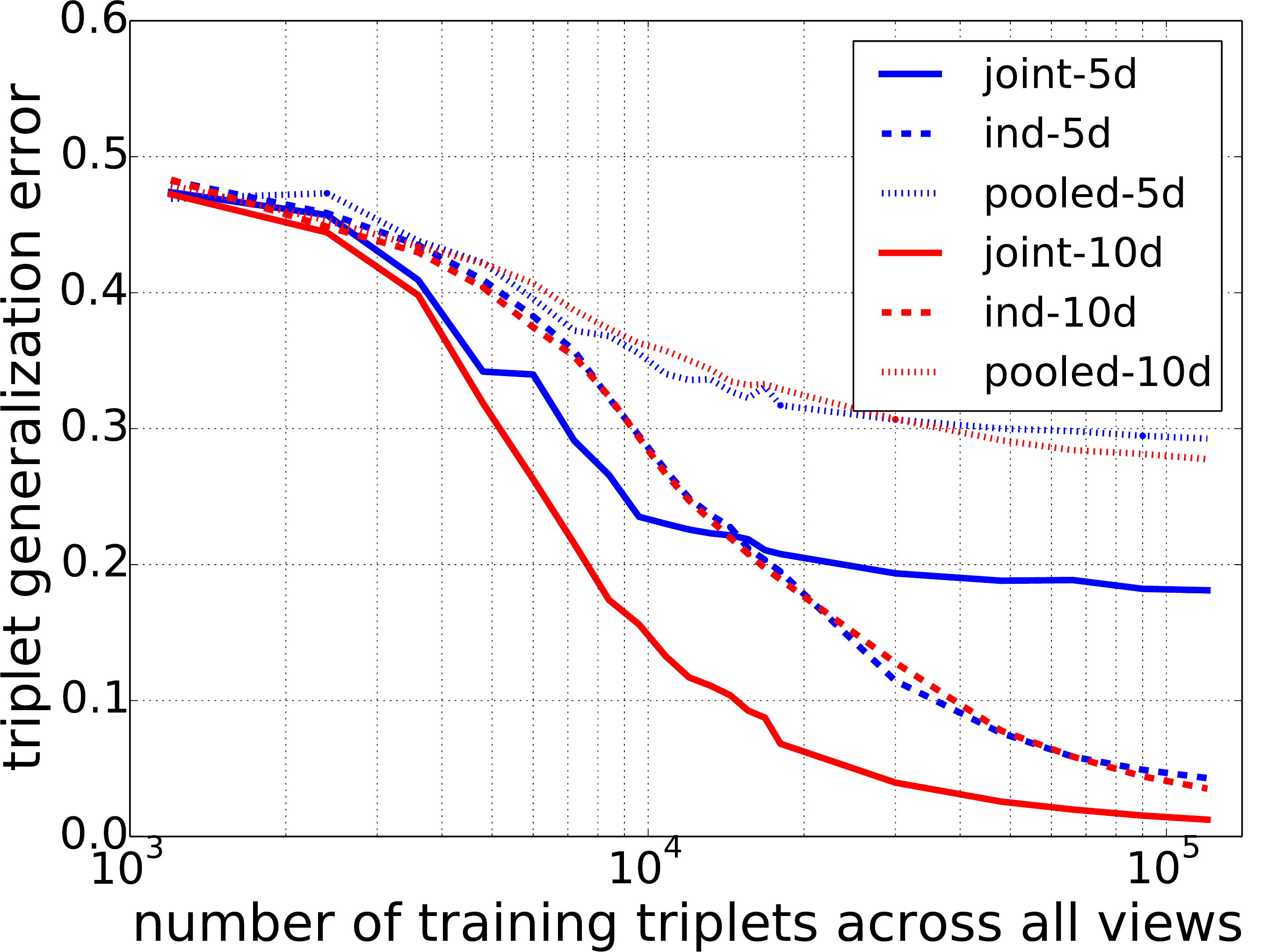}    
    }
	\subfigure[Poses of planes]{  
    \includegraphics[width=0.3\linewidth]{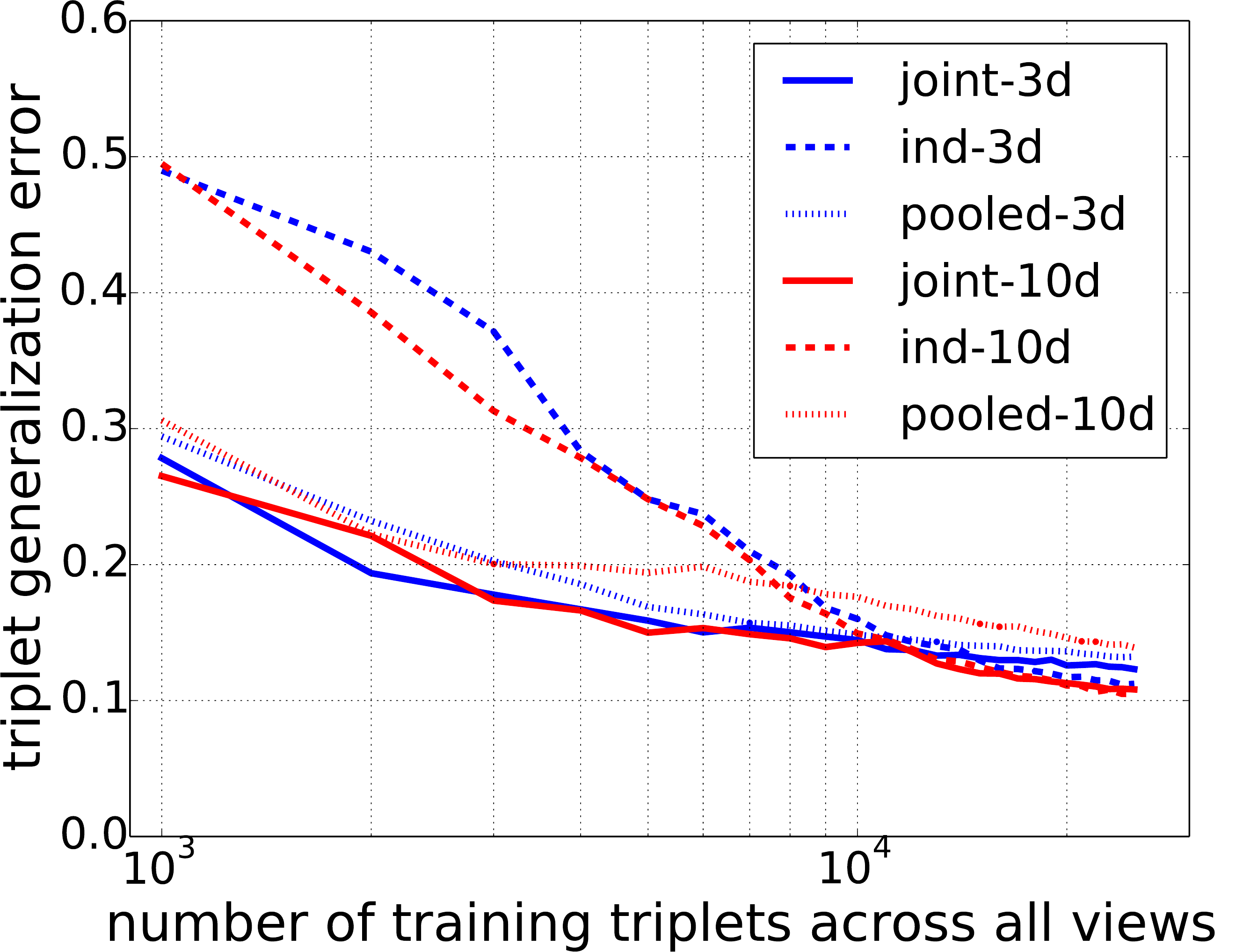}
    \label{fig:pose-results}
    }
    \caption{ Triplet generalization errors averaged
 across views for various datasets.}
 \label{fig:syn-and-pose-errors}
 \vspace{-0.15in}
\end{figure}

Embeddings were learned with embedding dimensions $D=5$ and 10.  Triplet generalization errors are plotted in
Fig.~\ref{fig:syn-and-pose-errors} (a) and (b) for clustered and uniform
data, respectively.  Our algorithm achieved lower triplet
generalization error than both {\bf independent} and {\bf pooled}
methods on both datasets. The improvement
was particularly large when the
number of triplets was limited (less than 10,000 for the clustered case).
The simple {\bf pooled} method was the worst on both datasets.
 Note that in contrast to the {\bf pooled} method, the proposed {\bf joint} method can
choose different embedding dimension 
 automatically  (due to the trace regularization)
for each view while maintaining a shared subspace.

\subsection{Poses of airplanes}

\paragraph{Description}
This dataset was constructed from 200 images of airplanes from the PASCAL
VOC dataset~\cite{everingham10pascal} which were annotated with 16
landmarks such as nose tip, wing tips, etc~\cite{boudev10detecting}. We
used these landmarks to construct a pose-based similarity. Given two
planes and the positions of landmarks in these images, pose similarity
was defined as the residual error of alignment between the two sets of
landmarks under scaling and translation. We generated 5 views each of
which was associated with a subset of these landmarks; see supplementary
material for details.  Three annotated
images from the set are shown in the left panel of
Fig.~\ref{fig:birds-views}.
 The planes are highly diverse ranging from passenger planes to fighter jets, varying in size and form which results in a slightly different similarity between instances for each view. However, there is a strong correlation between the views because the underlying set of landmarks are shared.


\paragraph{Results}
We used $D=3$ and $D=10$ as embedding dimensions.
Figure~\ref{fig:pose-results} shows the triplet generalization errors 
 of the three methods. The proposed joint
model performed clearly better than  {\bf independent}.
This was not only in average but also uniformly for each view (see
supplementary material). The {\bf pooled} method had a slightly larger
error than the proposed joint learning approach but better than the
{\bf independent} approach.

\subsection{CUB-200 birds data}

\paragraph{Description}
We used the dataset \cite{WelinderEtal2010} consisting of 200 species of
birds and use the annotations collected using the setup of Wah et
al. \shortcite{wah14similarity,wah15learning}. Briefly, similarity
triplets among images of each species were collected in a crowd-sourced
manner: every time, a user was asked to judge the similarity between an
image of a bird from the target specie $z_i$ and nine images of birds of
different species $\{ z_k \}_{k\in \mathcal{K}}$ using the interface of
Wilber et al.~\shortcite{wilber2014cost}, where $\mathcal{K}$ is the set of all 200 species.  For each display, the user partitioned these nine images into two sets, $\mathcal{K}_{sim}$ and $\mathcal{K}_{dissim}$, with $\mathcal{K}_{sim}$ containing birds considered similar to the target and $\mathcal{K}_{dissim}$ having the ones considered dissimilar.  Such a partition was broadcast to an equivalent set of triplet constraints on associated species, $\{(i,j,l) \mid j \in \mathcal{K}_{sim}, \, l \in \mathcal{K}_{dissim} \}$.  Therefore, for each user response, $\left|\mathcal{K}_{sim}\right|\left|\mathcal{K}_{dissim}\right|$ triplet constraints were obtained.

To collect view-specific triplets, we presented 5 different cropped
versions (\eg~beak,  breast, wing) of the bird images as shown in the right panel of Fig.~\ref{fig:birds-views} 
and used the same procedure as before to collect triplet comparisons.
We obtained about 100,000 triplets from the uncropped original images
and about 4,000 to 7,000 triplets from the 5 cropped views.
 This dataset reflects a more realistic situation where not all triplet relations are available and some of them may be noisy due to the nature of crowd-sourcing.

In addition to the triplet generalization error, 
we evaluated the embeddings in a classification
task using a biological taxonomy of the bird species. Note that
in Wah et al.~\shortcite{wah15learning} embeddings were used to
interactively categorize images; here we simplify this process
 to enable detailed comparisons. We manually grouped the 200 classes to
get 6 super classes so that the number of objects in all classes were
balanced. These class labels were not used in the training 
but allowed us to evaluate the quality of embeddings using
the leave-one-out (LOO) classification error. More precisely, at the test stage,
we predict the class label of each embedded point according to the 
labels of its 3-nearest-neighbours (3-NN) in the learned metric.

Finally, since more triplets were available from the first (uncropped) view
compared to other views, we first sampled equal numbers of
triplets in each view up to a total of 18,000 triplets. Afterwards, we
added triplets only to the first view.


\paragraph{Results}


\begin{figure*}
    \centering
	\subfigure[Triplet generalization error]{
    \includegraphics[width=0.3\linewidth]{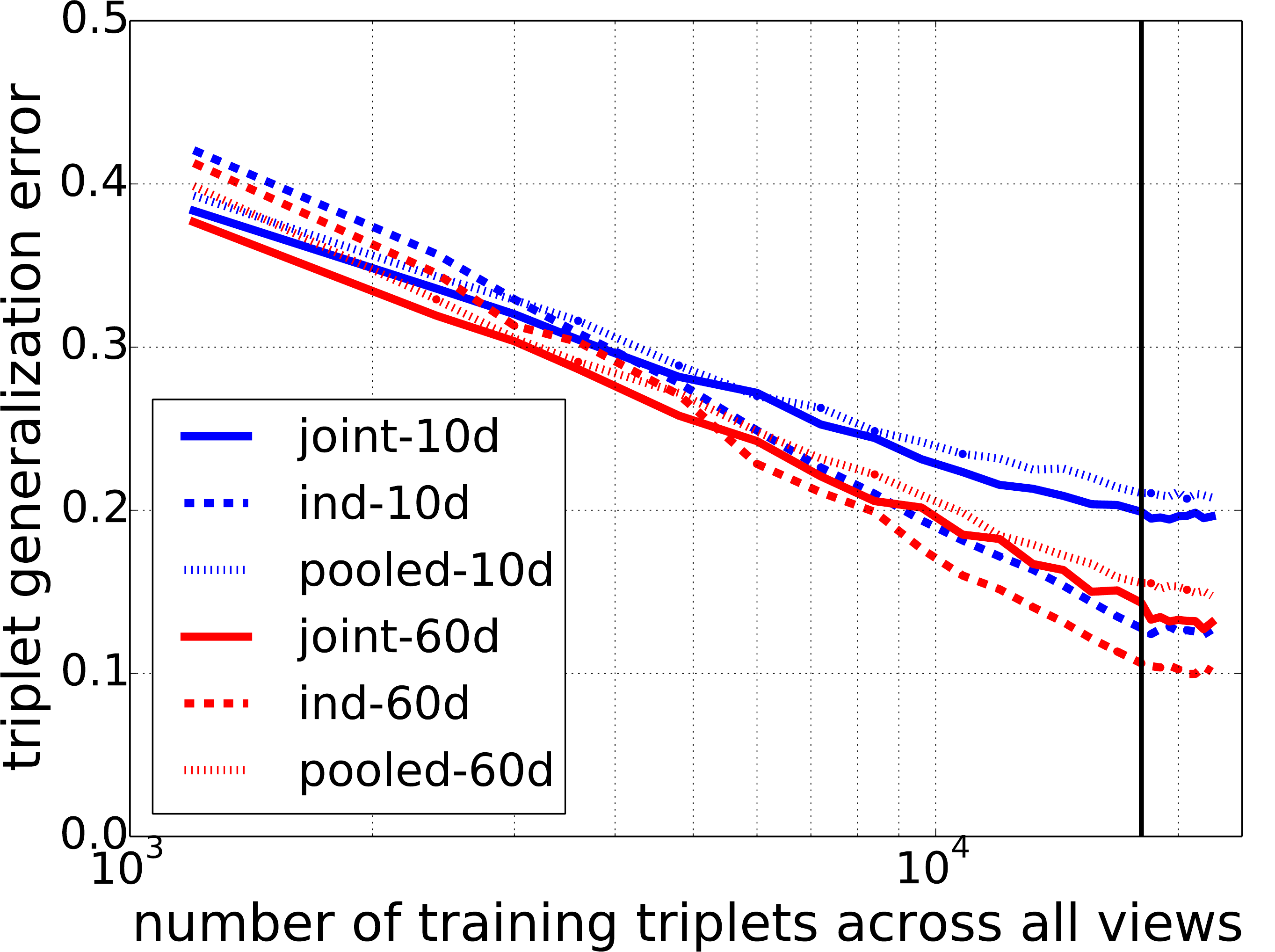}
    \label{fig:birds-results-gnmds-tripvio}
    }
    \subfigure[LOO 3-NN classification error]{
    \includegraphics[width=0.3\linewidth]{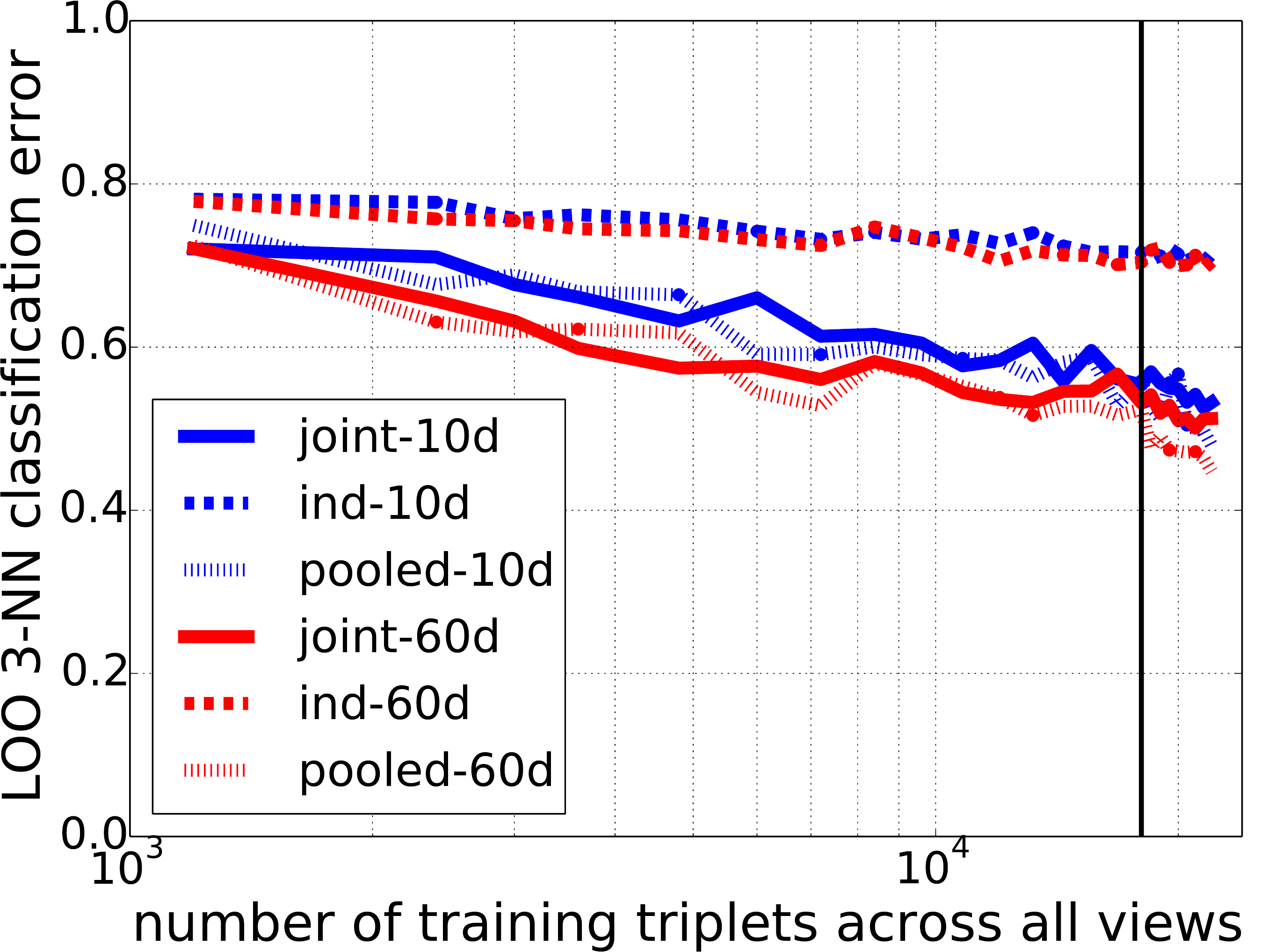}
    \label{fig:birds-results-gnmds-knn}
    }
    \subfigure[Learning a new view]{
    \includegraphics[width=0.3\linewidth]{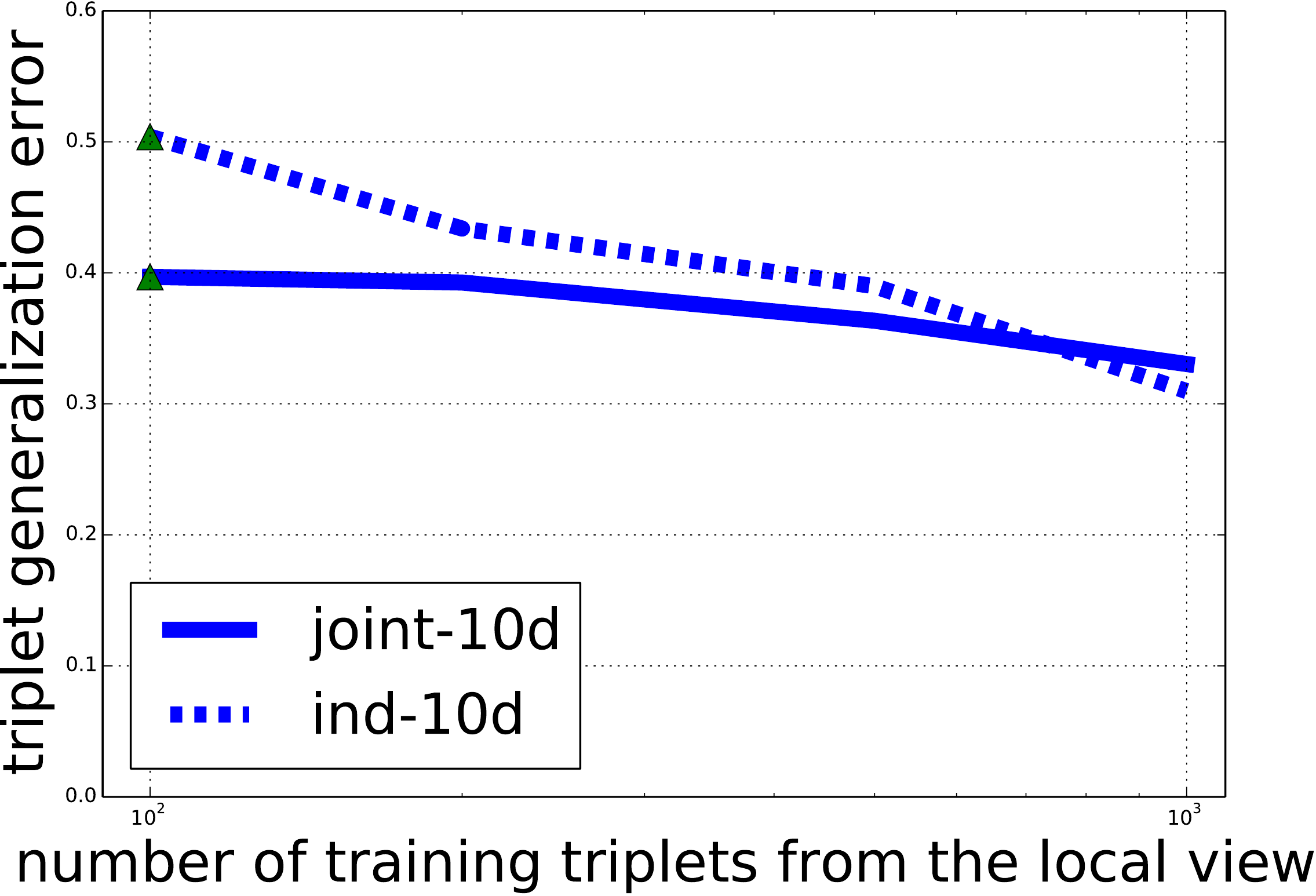}
    \label{fig:birds-emb-zeroshot}
    }
    \caption{ Results on CUB-200 birds dataset. (c) shows the triplet
 generalization error on the second view.}
    \label{fig:birds-results}
\end{figure*}


We used $D=10$ and $D=60$ as embedding dimensions; note that joint
learning in 60 dimensions roughly has the same number of parameters as
independent learning in 10 dimensions.
%
%
Figures \ref{fig:birds-results} (a) and (b) show the triplet generalization
errors and the LOO 3-NN classification errors, respectively.
The solid vertical line shows the point (18,000 triplets) that we start to add training
triplets only to the first view.
Comparing joint learning in 10 dimensions and 60 dimensions, we see that
the higher dimension gives the lower error.
The error of joint learning was better than independent
 learning for small number of triplets.
 Interestingly the error of joint learning in 60 dimensions
coincides with that of independent learning in 10 dimensions after
seeing 6,000 triplets. This can be explained by the fact that with 6
views, the two models have comparable complexity (see discussion at the
end of the previous section) and thus the same
asymptotic variance.
Our method obtains lower leave-one-out classification errors on all
views except for the first view; see supplementary material.

\paragraph{Learning a new view}
On the CUB-200 birds dataset, we simulated the situation of learning a new
view (or zero-shot learning).  We drew a training set that contains 100--1000
triplets from the second view and 3,000 triplets from all
other 5 views.  We investigated how joint learning helps in
estimating a good embedding on a new view with extremely small number of triplets.
The triplet generalization errors of both approaches 
are shown in Fig.~\ref{fig:birds-emb-zeroshot}.
 The triplet generalization error of the proposed joint
learning was lower than that of the independent learning up to around 700
triplets. The embedding of the second view learned jointly with other
views was clearly better than that learned independently and consistent
with the quantitative evaluation; see supplementary material.

\subsection{Performance gain and triplet consistency}
\label{sec:relation}
\begin{figure}[h!]
\begin{center}
\includegraphics[width=\linewidth]{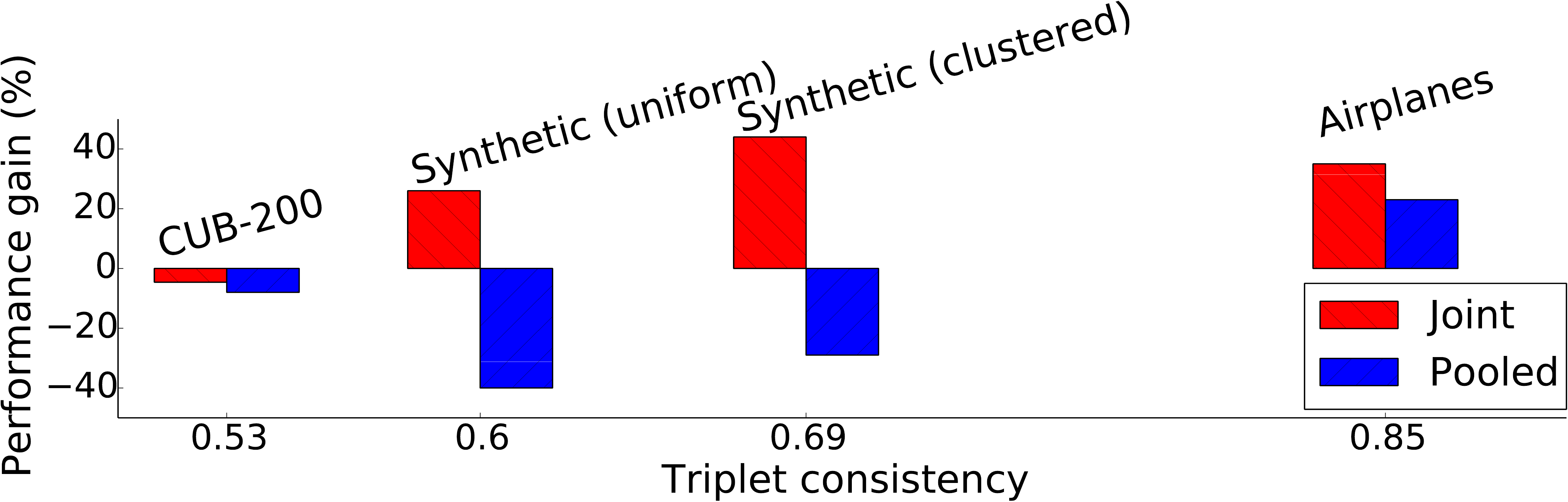}
\caption{\label{fig:gain} Relating the performance gains of joint and
 pooled learning with the triplet consistency. }
\end{center}
\end{figure}

In Fig.~\ref{fig:gain}, we relate the performance gain we obtained
for the joint/pooled learning approaches compared to the independent learning
approach with the underlying between-task similarity. The performance gain
was measured by the difference between the area under the triplet
generalization errors normalized by that of the independent learning.
The between-task similarity was measured by the triplet consistency between
two views averaged over all pairs of views. For the CUB-200 dataset
in which only a subset of valid triplet constraints are available, we
take the independently learned embeddings with the largest number of
triplets and use those to compute the triplet consistency. 

We can see that when the triplet consistency is very high, pooled
learning approach is good enough. However, when the triplet consistency
is not too high, it may harm to pool the triplets together. The proposed
joint learning approach has the most advantage in this intermediate regime.
On the other hand, the consistency was close to random (0.5) for the CUB-200 dataset possibly
explaining why the performance gain was not as significant as in the
other datasets.


\section{Incorporating features and class information}
\label{sec:incorp-feature}
The proposed method can be applied to a more general setting in which
each object comes with a feature vector and  a loss function not
derived from triplet constraints is used.

\begin{table*}[htb]
{\small 
\begin{center}
\caption{\label{tab:isolet}  Test error rates on ISOLET dataset. 
}
\begin{tabular}[tb] { p{0.5cm} |p{1.4cm}|p{1.4cm}|p{1.4cm}
||p{1.4cm}|p{1.4cm}|p{1.4cm} }
& \multicolumn{3}{c ||}{Tested with view-specific train data} & \multicolumn{3}{ c}{Tested with all train data}\\
\hline
Task & \multicolumn{1}{c |}{MT-LMNN} & \multicolumn{2}{c ||}{Proposed method} 
& \multicolumn{1}{c |}{MT-LMNN} & \multicolumn{2}{c}{Proposed method} \\
\hline
 & 378 dim & $D=169$ & $D=378$
 & 378 dim & $D=169$ & $D=378$\\
\hline
\hline
1 & 4.68 & {\bf 3.78} & 4.10 & 4.23 & {\bf 3.46} & 3.65 \\
2 & 4.55 & 3.91 & {\bf 3.52} & {\bf 3.14} & 3.84 & 3.40 \\
3 & 6.28 & {\bf 5.32} & 5.64 & 3.52 & {\bf 3.39} & 3.52 \\
4 & 7.76 & {\bf 5.83} & {\bf 5.83} & 4.23 & 4.10 & {\bf 3.52} \\
5 & 6.28 & {\bf 5.06} & 5.19 & 4.23 & {\bf 3.97} & {\bf 3.97} \\
\hline
Avg & 5.91 & {\bf 4.78} & 4.86 & 3.87 & 3.76 & {\bf 3.61} \\
\end{tabular}
\end{center}
}
\end{table*}
As an example, we employ the idea of multi-task large margin nearest
neighbor (MT-LMNN) algorithm \cite{parameswaran2010large} and adapt our model to handle a classification
problem. The loss function of MT-LMNN consists of 
two terms. The first term is a hinge loss for triplet constraints
as in \eqref{eq:multi-metric-obj} but the triplets are derived from
class labels. The second term is the sum of squared distances between each
object and its ``target neighbors'' which is also defined based on class labels;
see Weinberger et al. \shortcite{WeiBliSau06,parameswaran2010large} for details. 
The major difference between MT-LMNN
and our model is that MT-LMNN  parametrizes a local metric as the sum of a
global average $\bM_0$ and a view-specific metric $\bM_t$ as
$\bK_t=\bM_0+\bM_t$; thus the learned metric is generally full rank. On
the other hand, our method parametrizes it as a product of
global transform and local metric as $\bK_t=\bL\bM_t\bL^\top$, which
allows the local embedding dimension to be controlled by the trace regularization.

We conduct experiments on ISOLET spoken alphabet recognition dataset \cite{FanCol91} which consists 7797 examples of English alphabets spoken by 150 subjects and each example is described by a 617 dimension feature vector.  The task is to recognize the letter of each spoken example as one of the English alphabets.  The subjects are grouped into sets of 30 similar speakers leading to 5 tasks.  

We adapt the experimental setting from the work of MT-LMNN.  Data is first projected onto its first 378 leading
PCA components that capture 99 \% of variance. We train our model in a
$H=378$ dimensional space with $D=169$ and $378$, and compare it with
a MT-LMNN trained with the code provided by the authors.
In the experiment, each task is randomly divided into 60/20/20 subsets for 
train/validation/test.  We tuned the parameters on the validation sets.

Test error rates of 3-nearest-neighbor (3-NN) classifiers are reported in
Table~\ref{tab:isolet}. The left panel shows the errors using only the
view-specific training data for the classification. The right panel
shows those using all the training data with view-specific distance. Results are
averaged over 10 runs. Simpler baseline methods, such as, euclidean metric and
pooled (single task) learning are not included here because MT-LMNN already
performed better than them.
%
We can see that the proposed method performed better than MT-LMNN,
while learning in a 378 dimensional space and reducing to a 169 dimensional space
led to comparable error rates. A possible explanation for this mild
dependence on the choice of embedding dimension $D$ could be given by
the fact that both $\bL$ and $\bM_t$ are regularized and the effective
embedding dimension is determined by the regularization and not by the
choice of $D$; see Prop. \ref{lem:effective}.
 The averaged error rates reported in the original
paper using 169 PCA dimensions were 5.19 \% for the
view-specific case and 4.01 \% when all training data were used; our
numbers are still better than theirs.

\section{Related work\label{sec:relatedwork}}
Embedding of objects from triplet or paired distance
comparisons goes back to the work of Shepard \shortcite{She62a,She62b}
and Kruskal \shortcite{Kru64a,Kru64b} and studied 
extensively
\cite{agarwal2007generalized,tamuz2011adaptively,mcfee2009partial,mcfee2011learning,van2012stochastic}
recently. 


More recently, triplet embedding / metric learning problems that
involve multiple measures of similarity have been considered.
Parameswaran and Weinberger \shortcite{parameswaran2010large} 
aimed at jointly solving multiple related metric learning problems by
exploiting possible similarities. More specifically, they modeled
the inner product in each view by a {\em sum} of shared global matrix and a
view-specific local matrix. Moreover, Rai, Lian, and Carin \shortcite{RaiLiaCar14} proposed a
Bayesian approach to multi-task metric learning. 
Unfortunately, the sum structure in their
work typically do not produce a low-rank metric, which makes it
unsuitable for learning view-specific embeddings.
In contrast, our method models it as a {\em product}
of them allowing the trace norm regularizer to determine the rank of
each local metric.
Xie and Xing \shortcite{XieXin13} and Yu, Wang, and Tao \shortcite{YuWanTao12} studied metric learning problems with
multiple input views. This is different from our setting in which the
notion of similarity varies from view to view.
Amid and Ukkonen \shortcite{AmiUkk15} considered the task of multi-view triplet
embedding in which the view is a latent variable; they proposed a
greedy algorithm for finding the view membership of each object as well
as its embedding. It could be useful to combine this approach with ours
when we do not have enough resource to collect triplets from all possible views.

\section{Discussion}
\label{sec:discussion}

We have proposed a model for jointly learning multiple measures of
similarities from multi-view triplet observations. The proposed model
consists of a global transformation, which represents each object as a fixed
dimensional vector, and local view-specific metrics.

Experiments on both synthetic and real datasets have demonstrate that
our proposed joint model outperforms independent and
pooled learning approaches in most cases. Additionally, we have shown that
the advantage of our joint learning approach becomes the most prominent
when the views are similar but not too similar  (which can
be measured by triplet consistency). 
Morevoer, we have extended
our model to incorporate class labels and feature vectors. The
proposed model performed favorably compared to MT-LMNN on ISOLET dataset.
Since in many real applications,
similarity triplets can be expensive to obtain, jointly learning
similarity metrics is preferable as it can recover the underlying
structure using relatively small number of triplets.

One way to look at the proposed model is to view the shared global transformation
as controlling the complexity. However our experiments have shown that
generally the higher the dimension, the better the performance (except for
the ISOLET dataset tested with view-specific training data). Thus an alternative
explanation could be that the regularization on both the global
transformation $\bL$ and local metrics $\bM_t$ is implicitly controlling
the embedding dimension.

Future work includes extension of the current model to other loss
functions (e.g., the t-STE loss \cite{van2012stochastic}) and to the
setting in which we do not know which view each triplet came from.

\bibliographystyle{aaai}
\bibliography{reference}

\begin{thebibliography}{}

\bibitem[\protect\citeauthoryear{Agarwal \bgroup et al\mbox.\egroup
  }{2007}]{agarwal2007generalized}
Agarwal, S.; Wills, J.; Cayton, L.; Lanckriet, G.; Kriegman, D.~J.; and
  Belongie, S.
\newblock 2007.
\newblock Generalized non-metric multidimensional scaling.
\newblock In {\em International Conference on Artificial Intelligence and
  Statistics},  11--18.

\bibitem[\protect\citeauthoryear{Amid and Ukkonen}{2015}]{AmiUkk15}
Amid, E., and Ukkonen, A.
\newblock 2015.
\newblock Multiview triplet embedding: Learning attributes in multiple maps.
\newblock In {\em Proceedings of the 32nd International Conference on Machine
  Learning (ICML-15)},  1472--1480.

\bibitem[\protect\citeauthoryear{Bourdev \bgroup et al\mbox.\egroup
  }{2010}]{boudev10detecting}
Bourdev, L.; Maji, S.; Brox, T.; and Malik, J.
\newblock 2010.
\newblock Detecting people using mutually consistent poselet activations.
\newblock In {\em European Conference on Computer Vision (ECCV)}.

\bibitem[\protect\citeauthoryear{Chechik \bgroup et al\mbox.\egroup
  }{2010}]{CheShaShaBen10}
Chechik, G.; Sharma, V.; Shalit, U.; and Bengio, S.
\newblock 2010.
\newblock Large scale online learning of image similarity through ranking.
\newblock {\em J. Mach. Learn. Res.} 11:1109--1135.

\bibitem[\protect\citeauthoryear{Cox \bgroup et al\mbox.\egroup
  }{2000}]{CoxMilMinPapYia00}
Cox, I.~J.; Miller, M.~L.; Minka, T.~P.; Papathomas, T.~V.; and Yianilos, P.~N.
\newblock 2000.
\newblock The bayesian image retrieval system, pichunter: theory,
  implementation, and psychophysical experiments.
\newblock {\em Image Processing, IEEE Transactions on} 9(1):20--37.

\bibitem[\protect\citeauthoryear{Davis \bgroup et al\mbox.\egroup
  }{2007}]{DavKulJaiSraDhi07}
Davis, J.~V.; Kulis, B.; Jain, P.; Sra, S.; and Dhillon, I.~S.
\newblock 2007.
\newblock Information-theoretic metric learning.
\newblock In {\em Proceedings of the 24th international conference on Machine
  learning},  209--216.
\newblock ACM.

\bibitem[\protect\citeauthoryear{Everingham \bgroup et al\mbox.\egroup
  }{2010}]{everingham10pascal}
Everingham, M.; Van~Gool, L.; Williams, C. K.~I.; Winn, J.; and Zisserman, A.
\newblock 2010.
\newblock The pascal visual object classes (voc) challenge.
\newblock {\em International Journal of Computer Vision} 88(2):303--338.

\bibitem[\protect\citeauthoryear{Fanty and Cole}{1991}]{FanCol91}
Fanty, M.~A., and Cole, R.~A.
\newblock 1991.
\newblock Spoken letter recognition.
\newblock In {\em Adv. Neural. Inf. Process. Syst. 3},  220--226.

\bibitem[\protect\citeauthoryear{Fazel, Hindi, and Boyd}{2001}]{FazHinBoy01}
Fazel, M.; Hindi, H.; and Boyd, S.~P.
\newblock 2001.
\newblock {A Rank Minimization Heuristic with Application to Minimum Order
  System Approximation}.
\newblock In {\em {Proc. of the American Control Conference}}.

\bibitem[\protect\citeauthoryear{Horn and Johnson}{1991}]{HorJoh91}
Horn, R.~A., and Johnson, C.~R.
\newblock 1991.
\newblock {\em Topics in matrix analysis}.
\newblock Cambridge University Press.

\bibitem[\protect\citeauthoryear{Jamieson and Nowak}{2011}]{jamieson2011low}
Jamieson, K.~G., and Nowak, R.~D.
\newblock 2011.
\newblock Low-dimensional embedding using adaptively selected ordinal data.
\newblock In {\em Communication, Control, and Computing (Allerton), 2011 49th
  Annual Allerton Conference on},  1077--1084.
\newblock IEEE.

\bibitem[\protect\citeauthoryear{Kruskal}{1964a}]{Kru64a}
Kruskal, J.~B.
\newblock 1964a.
\newblock Multidimensional scaling by optimizing goodness of fit to a nonmetric
  hypothesis.
\newblock {\em Psychometrika} 29(1):1--27.

\bibitem[\protect\citeauthoryear{Kruskal}{1964b}]{Kru64b}
Kruskal, J.~B.
\newblock 1964b.
\newblock Nonmetric multidimensional scaling: a numerical method.
\newblock {\em Psychometrika} 29(2):115--129.

\bibitem[\protect\citeauthoryear{Kumar \bgroup et al\mbox.\egroup
  }{2009}]{kumar2009attribute}
Kumar, N.; Berg, A.~C.; Belhumeur, P.~N.; and Nayar, S.~K.
\newblock 2009.
\newblock Attribute and simile classifiers for face verification.
\newblock In {\em Computer Vision, 2009 IEEE 12th International Conference on},
   365--372.
\newblock IEEE.

\bibitem[\protect\citeauthoryear{McFee and Lanckriet}{2009}]{mcfee2009partial}
McFee, B., and Lanckriet, G.
\newblock 2009.
\newblock Partial order embedding with multiple kernels.
\newblock In {\em Proceedings of the 26th Annual International Conference on
  Machine Learning},  721--728.
\newblock ACM.

\bibitem[\protect\citeauthoryear{McFee and Lanckriet}{2011}]{mcfee2011learning}
McFee, B., and Lanckriet, G.
\newblock 2011.
\newblock Learning multi-modal similarity.
\newblock {\em The Journal of Machine Learning Research} 12:491--523.

\bibitem[\protect\citeauthoryear{Mikolov \bgroup et al\mbox.\egroup
  }{2013}]{mikolov2013efficient}
Mikolov, T.; Chen, K.; Corrado, G.; and Dean, J.
\newblock 2013.
\newblock Efficient estimation of word representations in vector space.
\newblock {\em arXiv preprint arXiv:1301.3781}.

\bibitem[\protect\citeauthoryear{Parameswaran and
  Weinberger}{2010}]{parameswaran2010large}
Parameswaran, S., and Weinberger, K.~Q.
\newblock 2010.
\newblock Large margin multi-task metric learning.
\newblock In {\em Advances in neural information processing systems},
  1867--1875.

\bibitem[\protect\citeauthoryear{Rai, Lian, and Carin}{2014}]{RaiLiaCar14}
Rai, P.; Lian, W.; and Carin, L.
\newblock 2014.
\newblock Bayesian multitask distance metric learning.
\newblock In {\em NIPS 2014 Workshop on Transfer and Multitask Learning}.

\bibitem[\protect\citeauthoryear{Shepard}{1962a}]{She62a}
Shepard, R.~N.
\newblock 1962a.
\newblock The analysis of proximities: Multidimensional scaling with an unknown
  distance function. {I.}
\newblock {\em Psychometrika} 27(2):125--140.

\bibitem[\protect\citeauthoryear{Shepard}{1962b}]{She62b}
Shepard, R.~N.
\newblock 1962b.
\newblock The analysis of proximities: Multidimensional scaling with an unknown
  distance function. {II}.
\newblock {\em Psychometrika} 27(3):219--246.

\bibitem[\protect\citeauthoryear{{Srebro}, {Rennie}, and
  {Jaakkola}}{2005}]{SreRenJaa05}
{Srebro}, N.; {Rennie}, J. D.~M.; and {Jaakkola}, T.~S.
\newblock 2005.
\newblock Maximum-margin matrix factorization.
\newblock In Saul, L.~K.; Weiss, Y.; and Bottou, L., eds., {\em Advances in
  NIPS 17}. Cambridge, MA: MIT Press.
\newblock  1329--1336.

\bibitem[\protect\citeauthoryear{Tamuz \bgroup et al\mbox.\egroup
  }{2011}]{tamuz2011adaptively}
Tamuz, O.; Liu, C.; Belongie, S.; Shamir, O.; and Kalai, A.~T.
\newblock 2011.
\newblock Adaptively learning the crowd kernel.
\newblock {\em arXiv preprint arXiv:1105.1033}.

\bibitem[\protect\citeauthoryear{van~der Maaten and
  Hinton}{2008}]{van2008visualizing}
van~der Maaten, L., and Hinton, G.
\newblock 2008.
\newblock Visualizing data using t-sne.
\newblock {\em Journal of Machine Learning Research} 9(2579-2605):85.

\bibitem[\protect\citeauthoryear{van~der Maaten and
  Weinberger}{2012}]{van2012stochastic}
van~der Maaten, L., and Weinberger, K.
\newblock 2012.
\newblock Stochastic triplet embedding.
\newblock In {\em Machine Learning for Signal Processing (MLSP), 2012 IEEE
  International Workshop on},  1--6.
\newblock IEEE.

\bibitem[\protect\citeauthoryear{Wah \bgroup et al\mbox.\egroup
  }{2014}]{wah14similarity}
Wah, C.; Horn, G.~V.; Branson, S.; Maji, S.; Perona, P.; and Belongie, S.
\newblock 2014.
\newblock Similarity comparisons for interactive fine-grained categorization.
\newblock In {\em Computer Vision and Pattern Recognition}.

\bibitem[\protect\citeauthoryear{Wah, Maji, and Belongie}{2015}]{wah15learning}
Wah, C.; Maji, S.; and Belongie, S.
\newblock 2015.
\newblock Learning localized perceptual similarity metrics for interactive
  categorization.
\newblock In {\em IEEE Winter Conference on Applications of Computer Vision,
  WACV}.

\bibitem[\protect\citeauthoryear{Weinberger, Blitzer, and
  Saul}{2006}]{WeiBliSau06}
Weinberger, K.~Q.; Blitzer, J.; and Saul, L.~K.
\newblock 2006.
\newblock Distance metric learning for large margin nearest neighbor
  classification.
\newblock In Weiss, Y.; Sch\"{o}lkopf, B.; and Platt, J., eds., {\em Adv.
  Neural. Inf. Process. Syst. 18}. MIT Press.
\newblock  1473--1480.

\bibitem[\protect\citeauthoryear{Welinder \bgroup et al\mbox.\egroup
  }{2010}]{WelinderEtal2010}
Welinder, P.; Branson, S.; Mita, T.; Wah, C.; Schroff, F.; Belongie, S.; and
  Perona, P.
\newblock 2010.
\newblock {Caltech-UCSD Birds 200}.
\newblock Technical Report CNS-TR-2010-001, California Institute of Technology.

\bibitem[\protect\citeauthoryear{Wilber, Kwak, and
  Belongie}{2014}]{wilber2014cost}
Wilber, M.~J.; Kwak, I.~S.; and Belongie, S.~J.
\newblock 2014.
\newblock Cost-effective hits for relative similarity comparisons.
\newblock {\em arXiv preprint arXiv:1404.3291}.

\bibitem[\protect\citeauthoryear{Xie and Xing}{2013}]{XieXin13}
Xie, P., and Xing, E.~P.
\newblock 2013.
\newblock Multi-modal distance metric learning.
\newblock In {\em Proceedings of the Twenty-Third international joint
  conference on Artificial Intelligence},  1806--1812.
\newblock AAAI Press.

\bibitem[\protect\citeauthoryear{Xing \bgroup et al\mbox.\egroup
  }{2002}]{xing2002distance}
Xing, E.~P.; Jordan, M.~I.; Russell, S.; and Ng, A.~Y.
\newblock 2002.
\newblock Distance metric learning with application to clustering with
  side-information.
\newblock In {\em Advances in neural information processing systems},
  505--512.

\bibitem[\protect\citeauthoryear{Yu, Wang, and Tao}{2012}]{YuWanTao12}
Yu, J.; Wang, M.; and Tao, D.
\newblock 2012.
\newblock Semisupervised multiview distance metric learning for cartoon
  synthesis.
\newblock {\em {IEEE} Trans. Image Process.} 21(11):4636--4648.

\end{thebibliography}

\newpage
\onecolumn
%

\begin{center}
{\bf\large Supplementary Material}
\end{center}
\vskip10pt

\section{Proof of Proposition 1}
We repeat the statement for convenience.
\begin{numberedprop}{1}
\begin{align*}
 \min_{\substack{\bL\in\mathbb{R}^{H\times D},\\
\bM_1,\ldots,\bM_T\in\mathbb{R}^{D\times D}} }
\left(\gamma\sum_{t=1}^{T}
{\rm tr}(\bM_t) + \beta \|\bL\|_F^2
:\, \bL\bM_t\bL^\top=\bK_t\, (\forall t)
\right)=
2\sqrt{
\beta\gamma}{\rm tr}\left(\sum\nolimits_{t=1}^{T}\bK_t\right)^{1/2}
\end{align*}
Here the power $1/2$ in the right-hand side is the matrix square root.
\end{numberedprop}

\begin{proof}
 Let's define $\bar{\bM}=\sum_{t=1}^{T}\bM_t$. For any decomposition
 $\bK_t=\bL\bM_t\bL^\top$, we have
\begin{align*}
 2\sqrt{\beta\gamma}{\rm tr}\left(\sum_{t=1}^{T}\bK_t\right)^{1/2}
&=2\sqrt{\beta\gamma}{\rm
 tr}\left(\bL\bar{\bM}\bL^\top\right)^{1/2}\\
&=2\sqrt{\beta\gamma}\|\bar{\bM}^{1/2}\bL\|_{\ast}\\
&=2\sqrt{\beta\gamma}\sum_{j=1}^{r}\sigma_j(\bar{\bM}^{1/2}\bL)\\
&\leq 2\sqrt{\beta\gamma}\sum_{j=1}^{r}\sigma_j(\bar{\bM}^{1/2})\sigma_j(\bL)\\
&\leq
 \sum_{j=1}^{r}\left(\gamma\sigma_j^2(\bar{\bM}^{1/2})+\beta\sigma_j^2(\bL)\right)\\
&=\gamma{\rm tr}(\bar{\bM})+\beta\|\bL\|_F^2,\\
&=\gamma\sum_{t=1}^{T}{\rm tr}(\bM_t)+\beta\|\bL\|_F^2
\end{align*}
where $\|\cdot\|_\ast$ is the nuclear norm \cite{FazHinBoy01}; the
 fourth line follows from Theorem 3.3.14 (a) in Horn \& Johnson
 \cite{HorJoh91}, and the fifth line is due to the arithmetic mean-geometric mean inequality.

Let $\bar{\bK}:=\sum_{t=1}^{T}\bK_t$ and $\bar{\bK}=\bU\bLambda\bU^\top$
be its eigenvalue decomposition.
The equality is achieved by choosing
\begin{align}
\label{eq:optL}
\bL&=\bU\bLambda^{1/4}(\gamma/\beta)^{1/4} \\
\label{eq:optMt}
\bM_t&=\bLambda^{-1/4}\bU^\top\bK_t\bU\bLambda^{-1/4}(\beta/\gamma)^{1/2}
 \quad (t=1,\ldots,T)
\end{align}
Note that even when $\bar{\bK}$ is
 singular, $\bK_t$ is spanned by $\bar{\bK}$ and by restricting to the
 subspace spanned by $\bar{\bK}$, the above discussion is still
 valid.
\end{proof}

This lemma can be understood analogously to the identity regarding the
nuclear norm\cite{SreRenJaa05}
\begin{align*}
 \|\bX\|_{\ast} =
 \min_{\bU,\bV}\frac{1}{2}\left(\|\bU\|_F^2+\|\bV\|_F^2\right)\quad
 \text{subject to} \quad\bX=\bU\bV^\top.
\end{align*}
Note that the fact that the ratio of the two hyperparameters
$\beta/\gamma$ can be absorbed in the scale ambiguity between $\bL$ and $\bM_t$
 as in \eqref{eq:optL} and \eqref{eq:optMt} is special to
multiplicative models like our model and the nuclear norm and would not hold for an additive model like MT-LMNN.

\clearpage
\section{Additional details and results}

\subsection{Synthetic dataset}

\begin{figure}
\centering
    \includegraphics[width=0.45\linewidth]{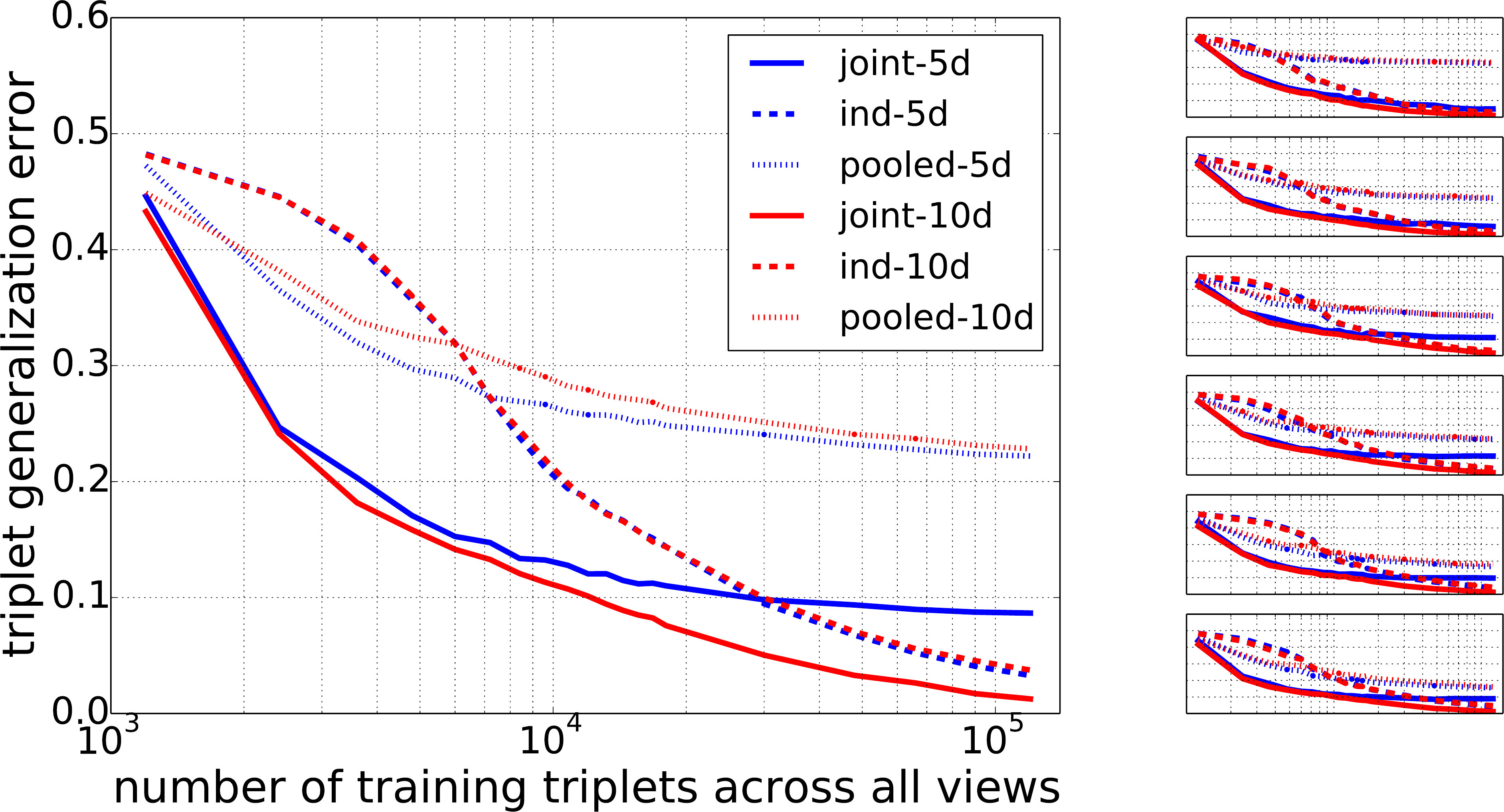}
    ~
    \includegraphics[width=0.45\linewidth]{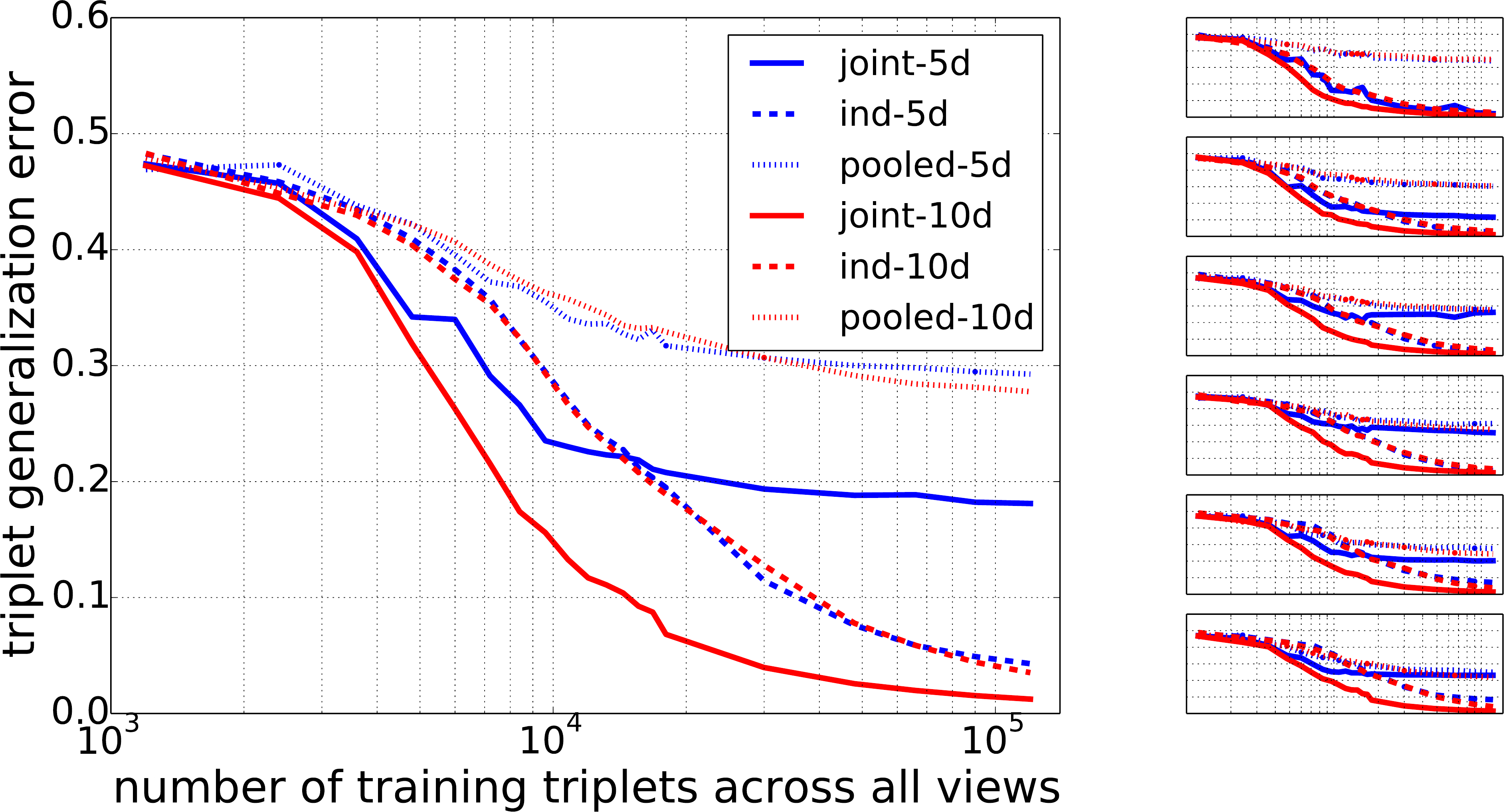}
    \caption{ Triplet generalization errors. The small figures shows errors on individual \emph{views} and the large figures show the average. (Left) Clustered synthetic data. (Right) Uniformly distributed data. }
    \label{fig:syn-errors}
\end{figure}

\begin{figure}
\centering
    \includegraphics[width=0.45\linewidth]{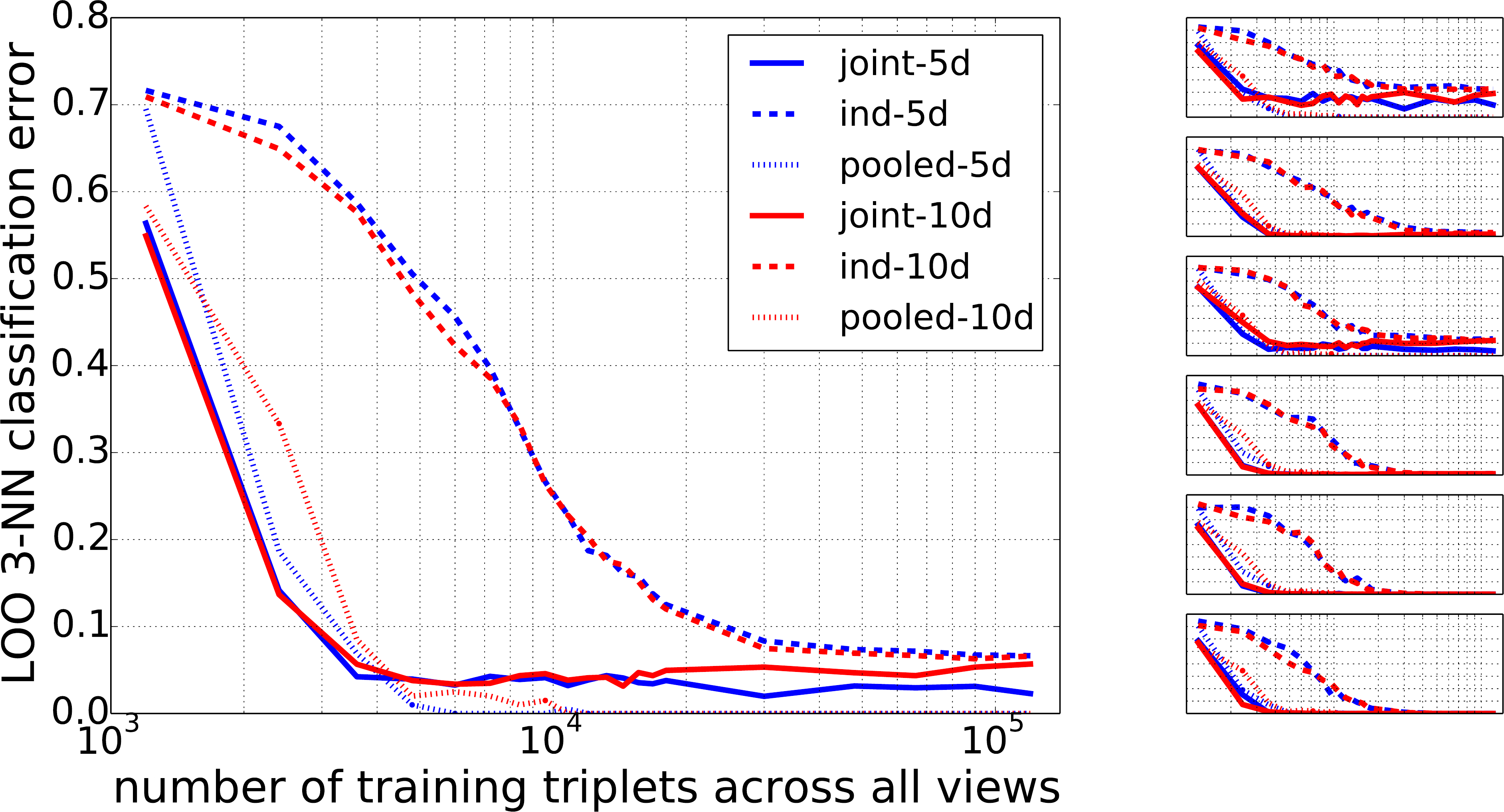}
    \caption{ Leave-one-out 3-nearest-neighbour classification errors on clustered synthetic data. The small figures shows errors on individual \emph{views} and the large figures show the average.}
    \label{fig:syn-loo-errors}
\end{figure}

In addition to the results in main paper, we illustrate view-specific triplet generalization error in Figure~\ref{fig:syn-errors} and leave-one-out classification error for clustered synthetic data in Figure~\ref{fig:syn-loo-errors}.

\subsection{Poses of airplanes dataset}
\subsubsection*{Details of annotations and view generation}
Each of the 200 airplanes were annotated with 16 landmarks namely, \\
\begin{table}[h]
\centering
\begin{tabular}{llll}
01. Top\_Rudder & 05. L\_WingTip &     09. Nose\_Bottom & 13. Left\_Engine\_Back \\
 02. Bot\_Rudder  &06. R\_WingTip &     10. Left\_Wing\_Base & 14. Right\_Engine\_Front \\
    03. L\_Stabilizer & 07. NoseTip &    11. Right\_Wing\_Base  &    15. Right\_Engine\_Back \\
    04. R\_Stabilizer &  08. Nose\_Top &    12. Left\_Engine\_Front & 16. Bot\_Rudder\_Front\\

\end{tabular}
\end{table}

\begin{figure}[h]
\centering
\includegraphics[width=0.7\linewidth]{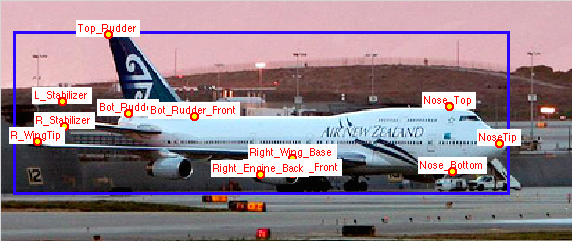}\\
\includegraphics[width=0.7\linewidth]{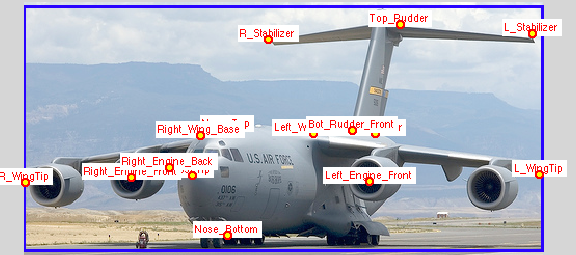}\\
\includegraphics[width=0.7\linewidth]{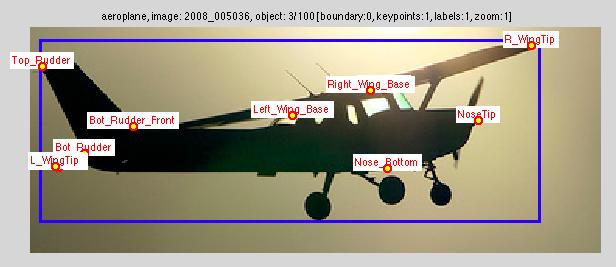}\\
\caption{Landmarks illustrated on the several planes}\label{f:landmarks}
\end{figure}

This is also illustrated in the Figure~\ref{f:landmarks}. The five different views are defined by considering different subsets of landmarks as follows:
\begin{enumerate}
\item \emph{all} $ \in \{1,2, \ldots, 20\}$
\item \emph{back} $ \in \{1,2, 3, 4, 16\}$
\item \emph{nose} $ \in \{7,8,9\}$
\item \emph{back+wings} $\in\{1, 2, \ldots, 6, 10, 11, \ldots, 16\}$
\item \emph{nose+wings} $ \in \{5, 6, \ldots, 15\}$
\end{enumerate}

For triplet $(A, B, C)$ we compute similarity $s_i(A,B)$ and $s_i(A, C)$ by aligning the subset $i$ of landmarks of $B$ and $C$ to $A$ under a translation and scaling that minimizes the sum of squared error after alignment. The similarity is inversely proportional to the residual error. This is also known as ``procrustes analysis'' commonly used for matching shapes.

\begin{figure}
\centering
	\subfigure{
    \includegraphics[width=0.45\linewidth]{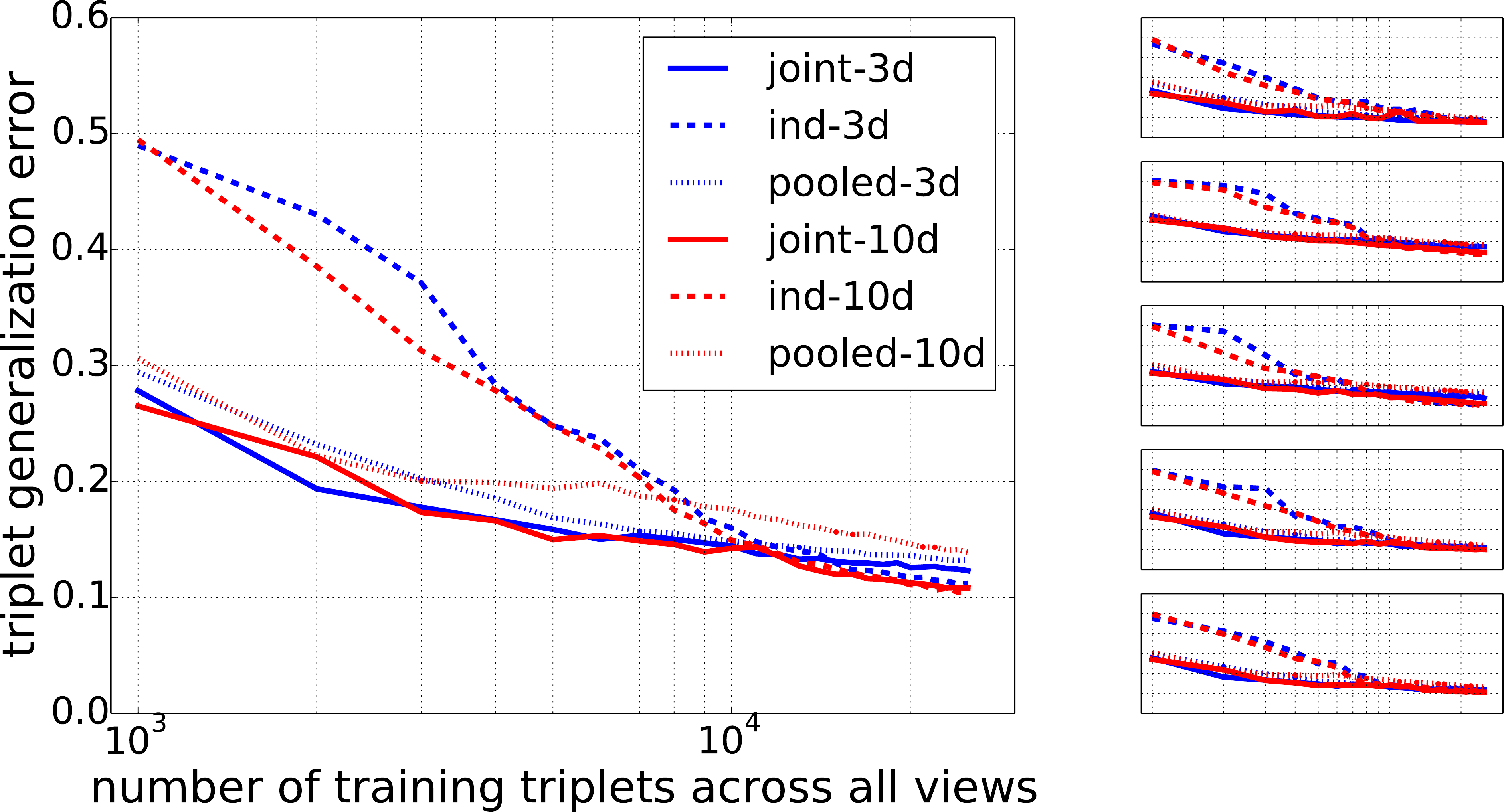}
    }
    ~
    \subfigure{
    \includegraphics[width=0.45\linewidth]{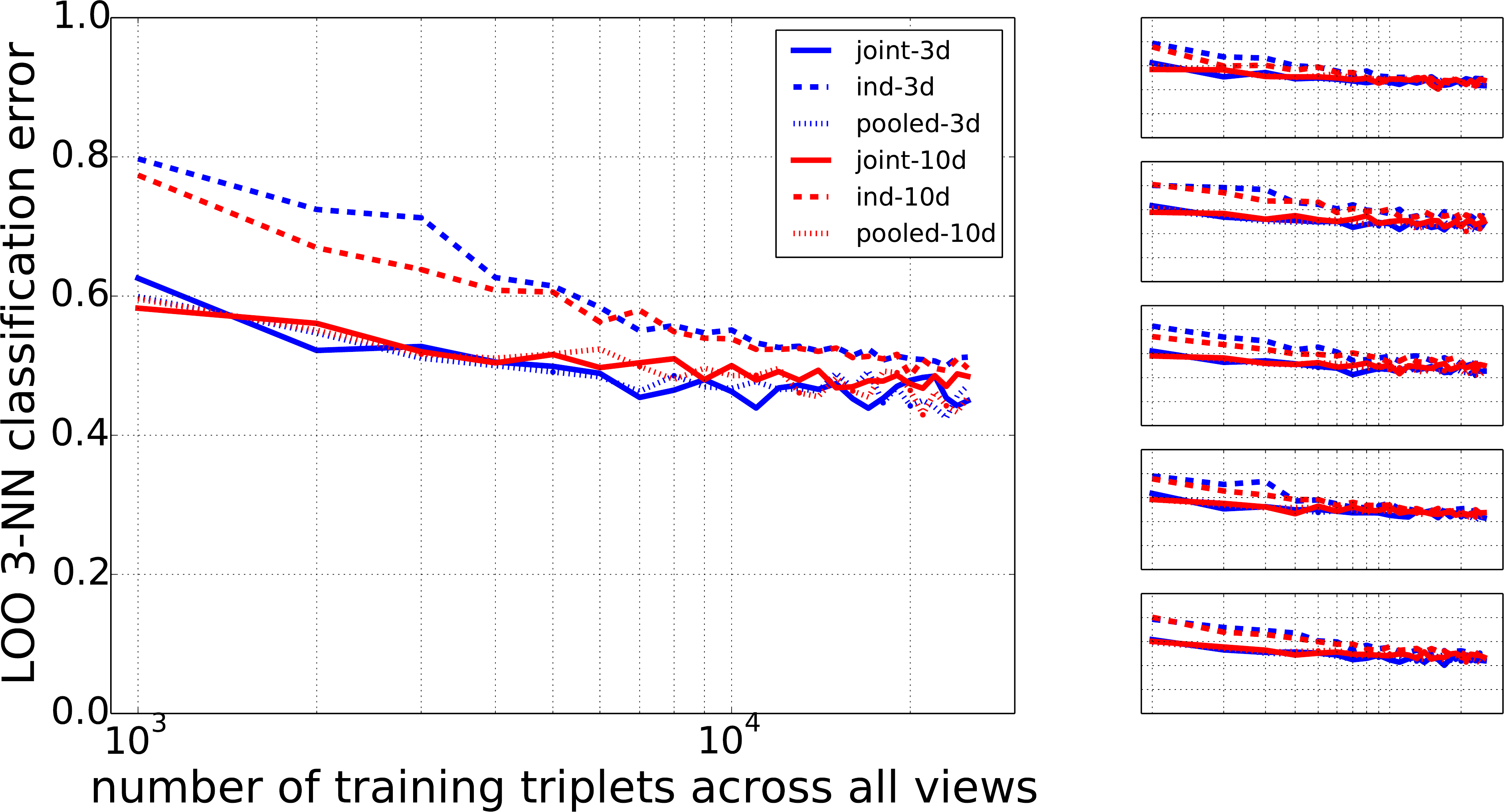}
    }
    
    \caption{ Experimental results on poses of planes dataset.  The small figures shows errors on individual \emph{views} and the large figures show the average.  (Left) Triplet generalization errors on poses of planes dataset. (Right) Leave-one-out 3-nearest-neighbour classification error.}
    \label{fig:pose-results-appendix}
\end{figure}

In addition to the results in main paper, we illustrate view-specific triplet generalization error and leave-one-out 3-nearest-neighbour classification error in Figure~\ref{fig:pose-results-appendix}.

\subsubsection*{Learned embedding}
\begin{figure}
    \centering
    \includegraphics[width=0.8\columnwidth]{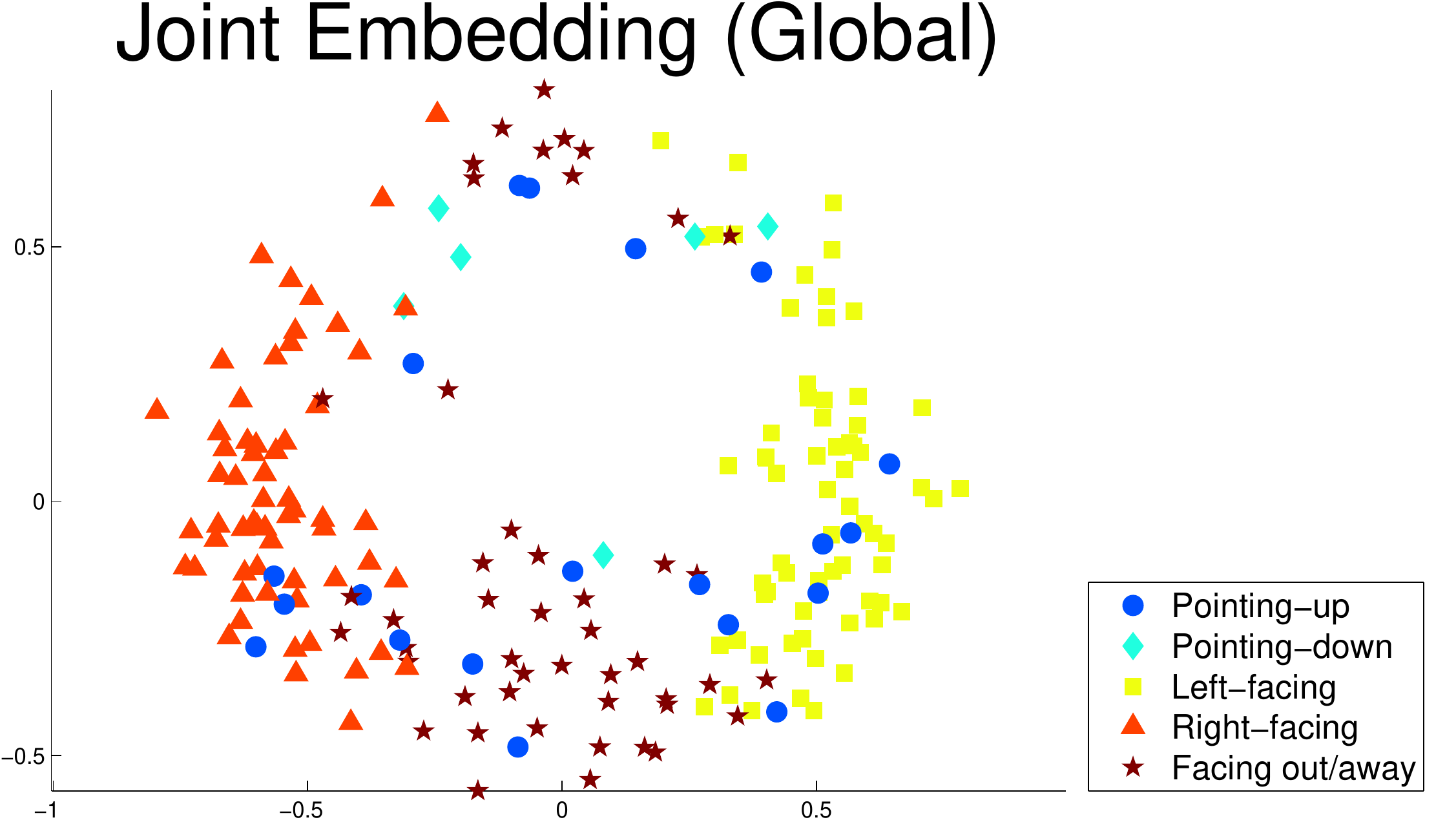}
    \caption{ The \emph{global view} of embeddings of poses of planes. \label{fig:pose-embd}}
\end{figure}

Figure \ref{fig:pose-embd} shows a 2D projection of the global view of
the objects onto their first two principle dimensions.
The visualization shows that objects roughly lies on a circle
 corresponding to the left-right and up-down orientation.

\subsection{CUB-200 birds dataset}

\begin{figure*}
    \centering
	\subfigure[]{
    \includegraphics[width=0.45\linewidth]{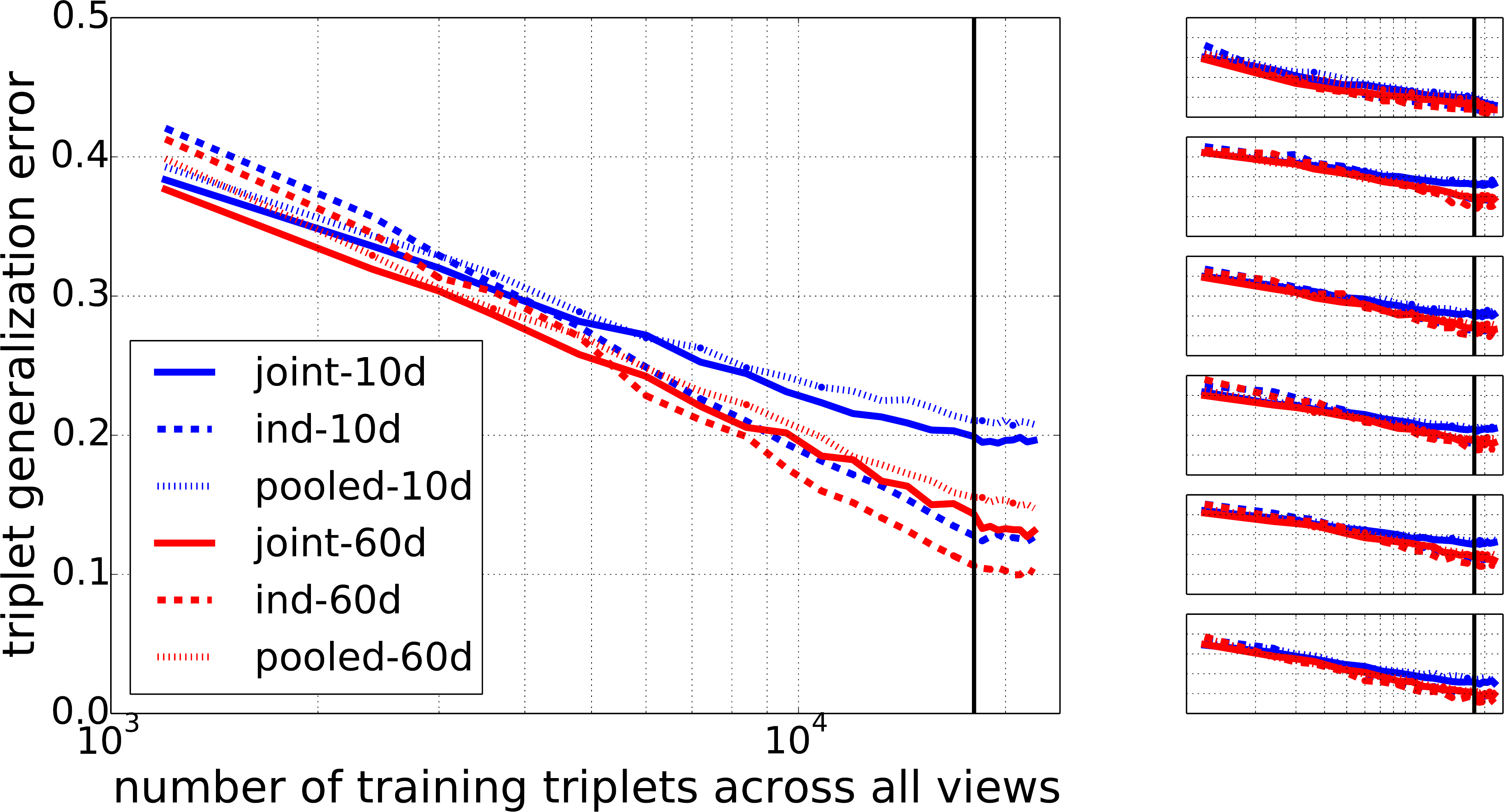}
    \label{fig:birds-results-gnmds-tripvio-appendix}
    }
    \subfigure[]{
    \includegraphics[width=0.45\linewidth]{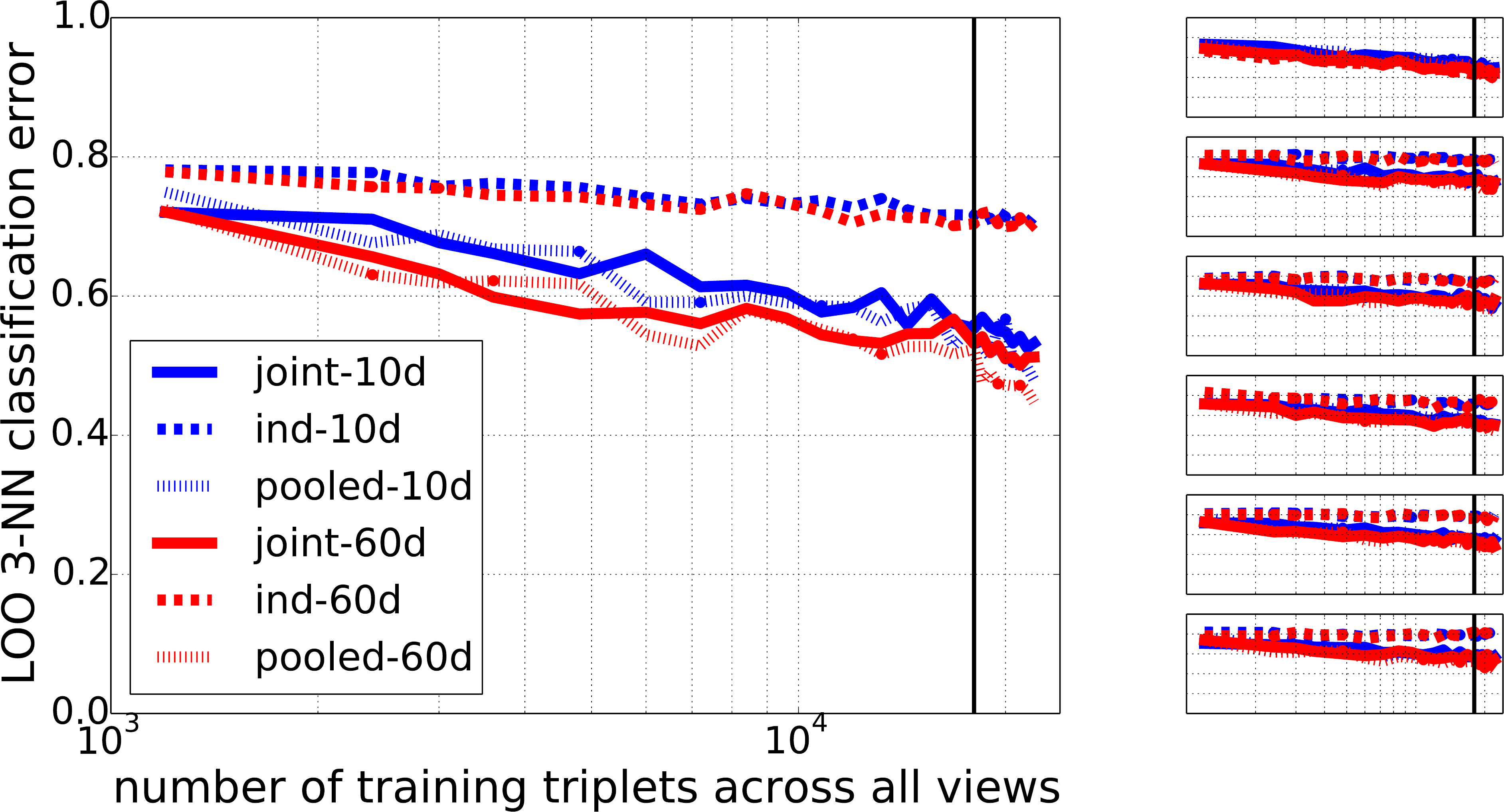}
    \label{fig:birds-results-gnmds-knn-appendix}
    }
    \caption{ Results on CUB-200 birds dataset. The small figures shows errors on individual \emph{views} and the large figures show the average. (a) triplet generalization error. (b) leave-one-out 3-nearest-neighbor classification error. }
    \label{fig:birds-results-appendix}
\end{figure*}

Here, we also include the view-specific generalization errors and leave-one-out classification errors for CUB-200 Birds Dataset.  See Figure~\ref{fig:birds-results-appendix}.

\subsection{Public figures face dataset}

\subsubsection*{Description}
Public Figures Face Database is created by Kumar \etal
\cite{kumar2009attribute}.  It consists of 58,797 images of 200 people.
Every image is characterized by 75 attributes which are real valued and
describe the appearance of the person in the image.  We selected 39 of
the attributes and categorized them into 5 groups according to the
aspects they describe: \emph{hair, age, accessory, shape} and
\emph{ethnicity}.  We randomly selected ten people and drew 20 images
for each of them to create a dataset with 200 images.  Similarity
between instances for a given group is equal to the dot product between
their attribute vectors where the attributes are restricted to those in
the group. We describe the details of these attributes below. Each group is considered as a \textit{local view} and \emph{identities} of the people in the images are considered as class labels.

\subsubsection*{Attributes}
\label{sec:attributes}
Each image in the \emph{Public Figures Face Dataset (Pubfig)} \footnote{Available at \url{http://www.cs.columbia.edu/CAVE/databases/pubfig/}} is characterized by 75 attributes.  We used 39 of the attributes in our work and categorized them into 5 groups according to the aspects they describe.  Here is a table of the categories and attributes:

\begin{center}
\begin{table}[h]
\begin{tabular}{ | l | p{13cm} |}
  \hline
  Category & Attributes\\
  \hline
  Hair & \emph{Black Hair, Blond Hair, Brown Hair, Gray Hair, Bald, Curly Hair, Wavy Hair, Straight Hair, Receding Hairline, Bangs, Sideburns.} \\
  \hline
  Age &  \emph{Baby,Child,Youth,Middle Aged,Senior.} \\
  \hline
  Accessory & \emph{No Eyewear, Eyeglasses, Sunglasses, Wearing Hat, Wearing Lipstick, Heavy Makeup, Wearing Earrings, Wearing Necktie, Wearing Necklace.} \\
  \hline
  Shape & \emph{Oval Face, Round Face, Square Face, High Cheekbones, Big Nose, Pointy Nose, Round Jaw, Narrow Eyes, Big Lips, Strong Nose-Mouth Lines.} \\
  \hline
  Ethnicity & \emph{Asian, Black, White, Indian.} \\
  \hline
\end{tabular}
\caption{List of Pubfig attributes that were used in our work.}
\end{table}
\end{center}

\subsubsection*{Results}

\begin{figure*}
    \centering
    \subfigure[]{
    \includegraphics[width=0.8\columnwidth]{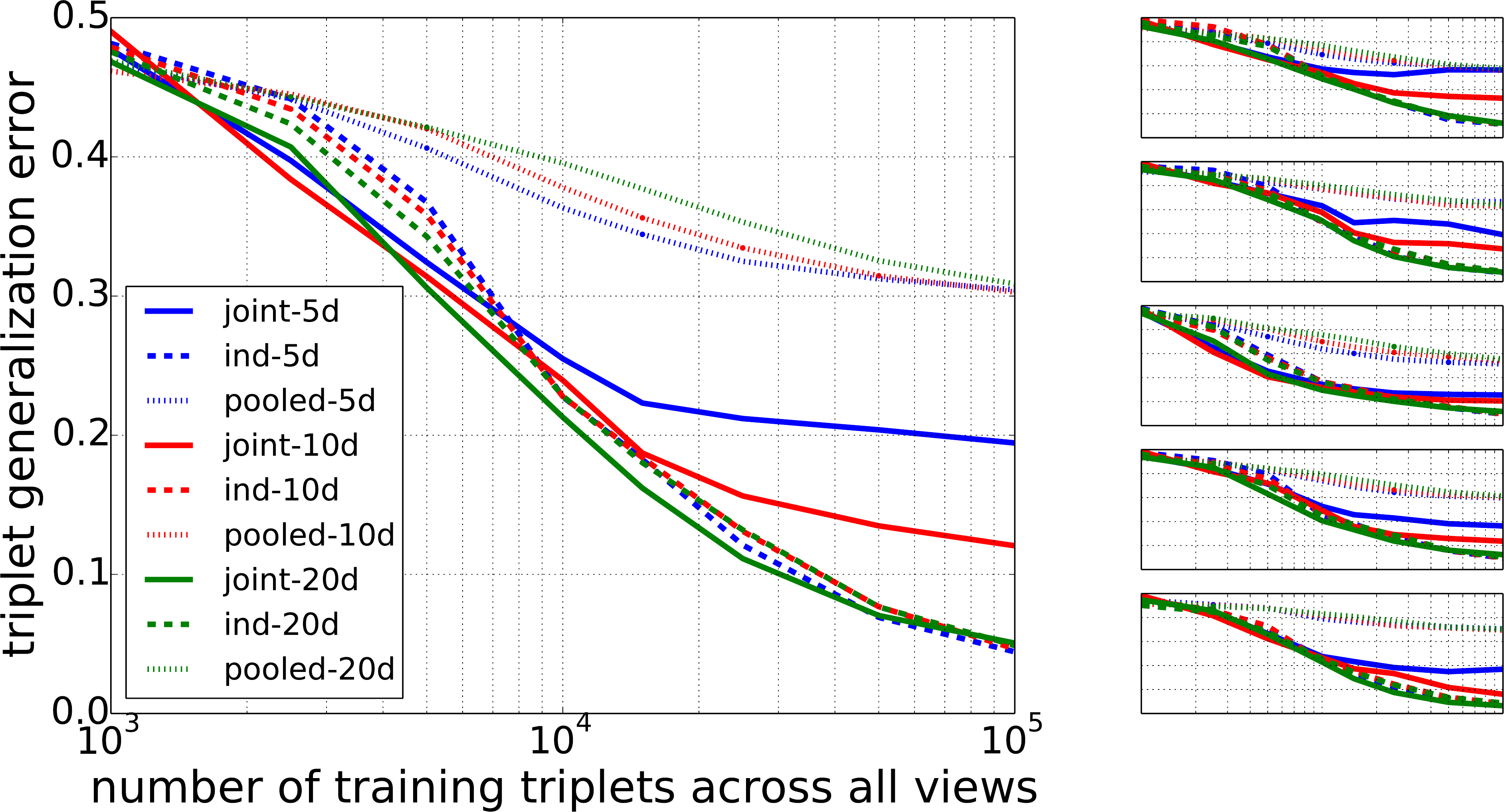}
    }
    \subfigure[]{
    \includegraphics[width=0.8\columnwidth]{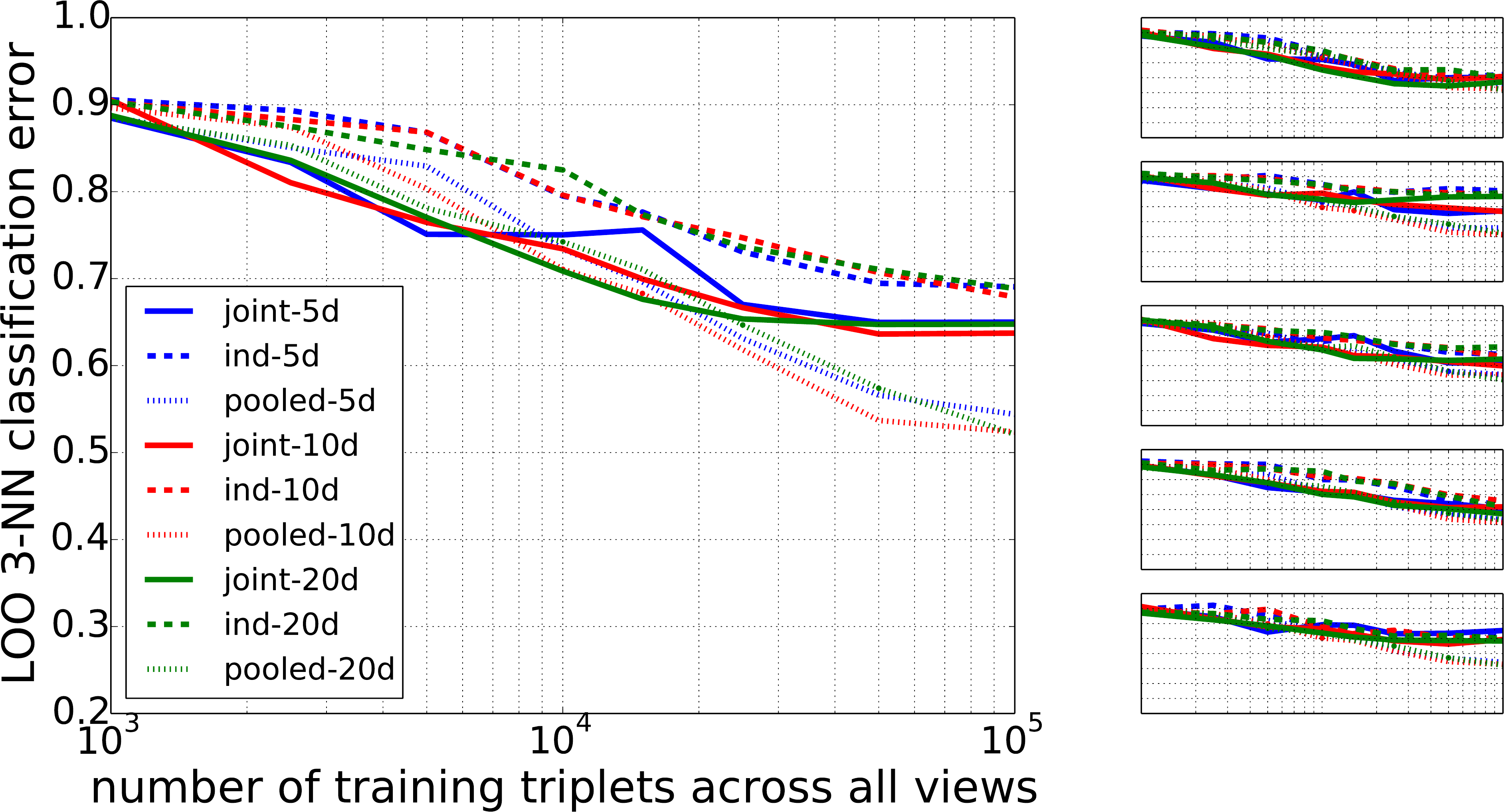}
    }
    \caption{ Results on public figures faces dataset.  Embeddings are learned in a 5 dimensional space, a 10 dimensional space and a 20 dimensional space.  (a) Triplet generalization error.  (b) Leave-one-out 3-nearest-neighbor classification error.   The small figures shows errors on individual \emph{views} and the large figures show the average. \label{fig:pubfig-results} }
\end{figure*}

The 200 images are embedded into 5, 10, and 20 dimensional spaces.  We
draw triplets randomly from the ground truth similarity measure to form
training and test sets.
 Triplet generalization errors and classification errors are
shown in Fig.~\ref{fig:pubfig-results}. 

In terms of the triplet generalization error, the joint learning reduces
the error faster than the independent learning up to around 10,000
triplets where the decrease slows down. Since the error in this regime
reduces monotonically with increasing number of dimensions, this can be
understood as a {\em bias} induced by the joint learning. On the other hand,
when we have less than 10,000 triples, the error of the joint learning
increases (but not as large as the independent learning) as dimension
increases; this can be understood as a {\em variance}.
When embedding in a 20 dimensional space, the joint learning 
has lower or comparable error to independent learning
even when $10^5$ triplets are available.
In terms of the leave-one-out classification error, joint learning
continues to be better even when the number of triplets are very large.


\subsection{Learning a new view}
\begin{figure}[tb]
    \centering
    \includegraphics[width=0.3\textwidth]{figures/birds/birds-zeroshot-tripvio.pdf}
    \includegraphics[width=0.3\textwidth]{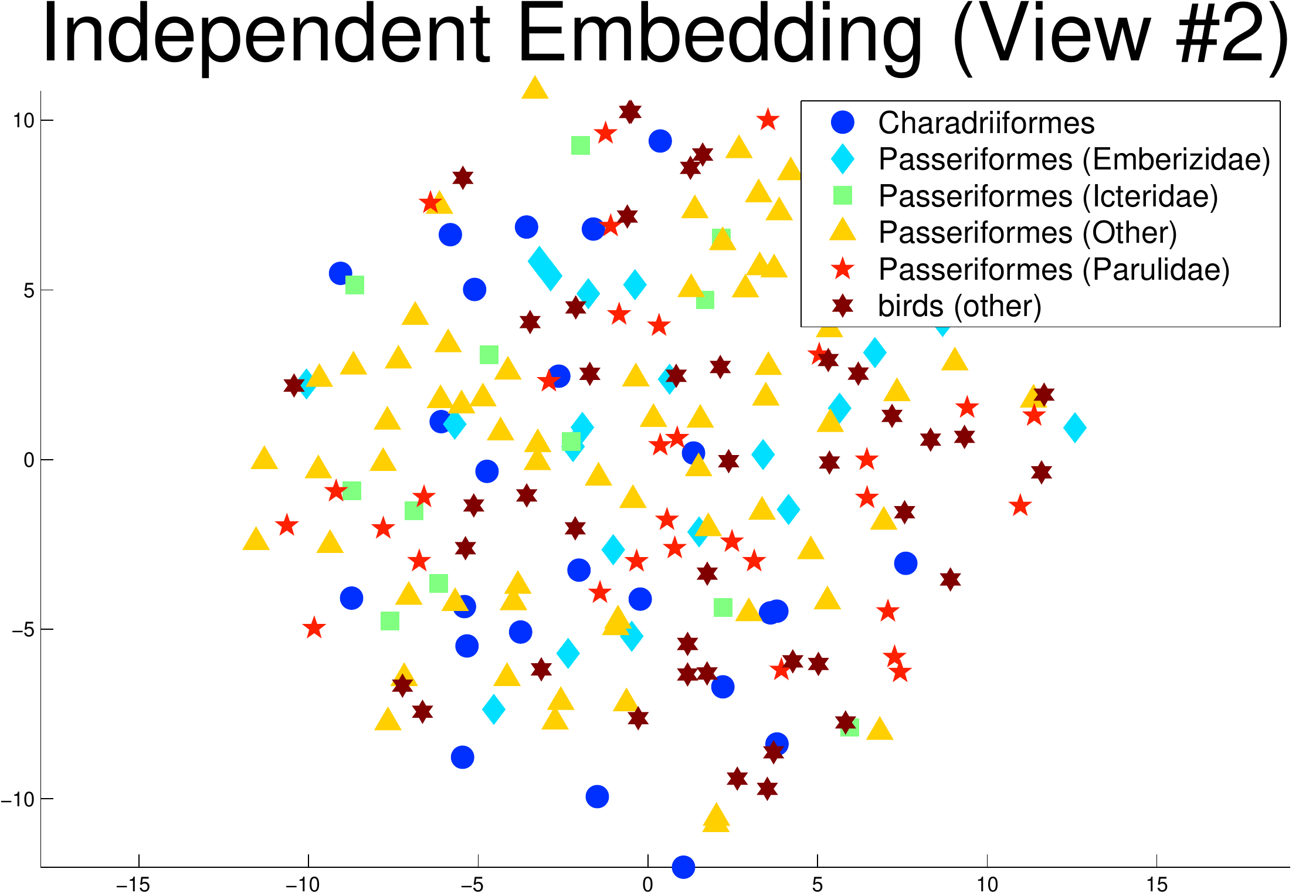}
    \includegraphics[width=0.3\textwidth]{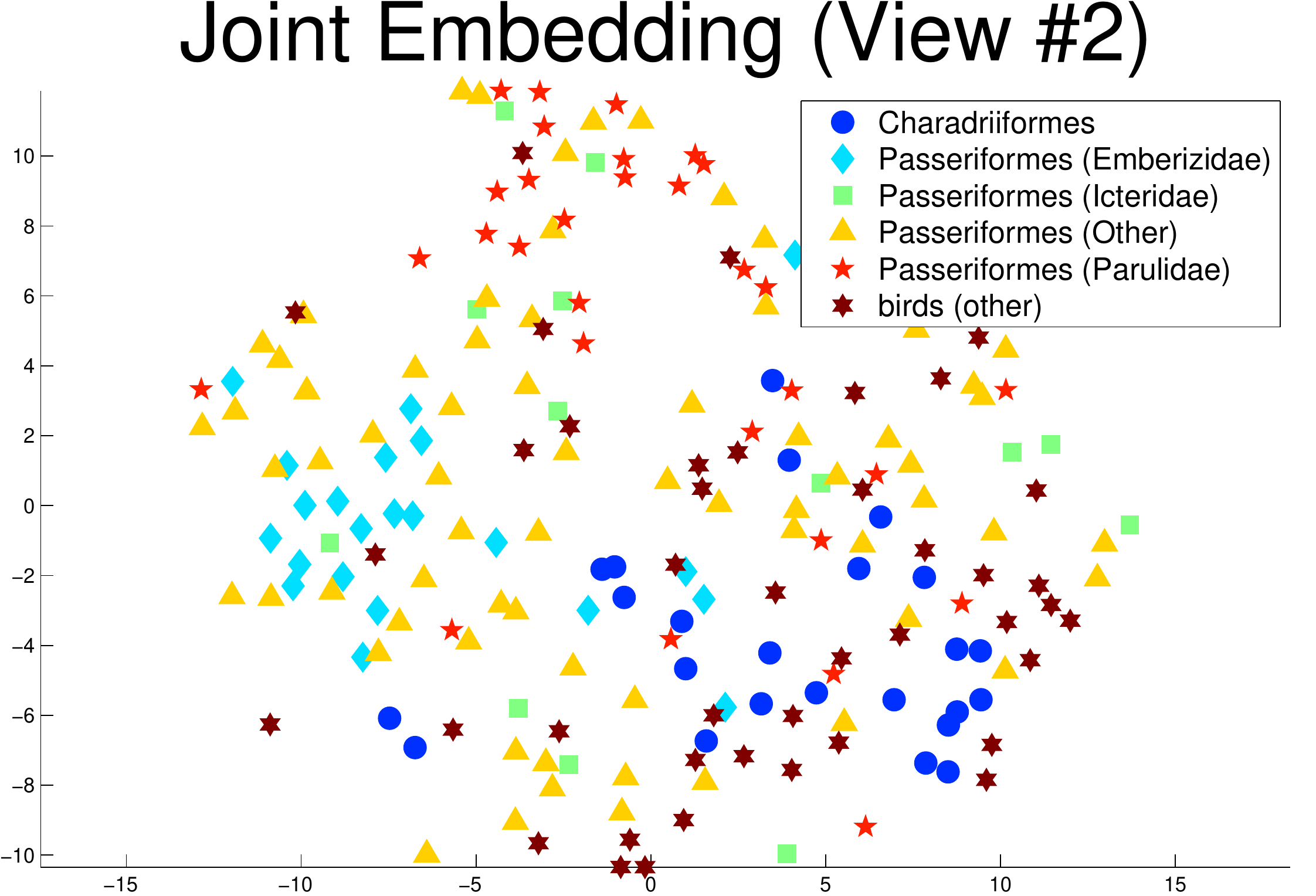}
    \caption{Learning a new view on CUB-200 birds dataset. Training data contains 100 triplets from the second \emph{local view} and 3,000 triplets from other 5 \emph{views}. Embeddings are learned in a 10 dimensional space and then further embedded in a 2 dimensional plane by using tSNE \cite{van2008visualizing} for the purpose of visualization.  Left: triplet generalization error on the second \emph{local view}.  Middle: embedding learned independently.  Right: embedding learned jointly.   \label{fig:birds-emb-zeroshot-appendix}}
\end{figure}

Figure \ref{fig:birds-emb-zeroshot-appendix} shows a 2D projection of the
embeddings learned by the {\bf independent} approach and the proposed
{\bf joint} approach in the setting for CUB-200 birds dataset described in the part of {\bf ``learning a new view''} in the main text.
Clearly the proposed joint learning
approach obtains a better separated clusters compared to the independent approach.

\subsection{Relating the performance gain with the triplet consistency}
\begin{table}[h]
 \begin{center}
\caption{Relating the performance gain of joint and pooled learning with the
  between-task similarity.}
  \label{tab:relation}
{\small   \begin{tabular}[tb]{p{2.5cm}|p{1.8cm}|p{1.8cm}|p{1.8cm}|p{1.8cm}|p{1.8cm}}
&  CUB-200 & PubFig & Synthetic (uniform) & Synthetic (clustered) & Airplanes  \\
\hline\hline
Average triplet consistency &
0.53 & 0.59 & 0.6 & 0.69  & 0.85 \\
\hline
Performance gain of joint learning(\%) &
-4.6 & 6.5 & 26 & 44  & 35 \\
\hline
Performance gain of pooled learning(\%) &
-8.0 & -56 & -40 & -29 & 23
  \end{tabular}}
 \end{center}
\end{table}


\end{document}